\documentclass[10pt,twocolumn,letterpaper]{article}

\usepackage{iccv}
\usepackage{times}
\usepackage{epsfig}
\usepackage{graphicx}
\usepackage{amsmath}
\usepackage{amssymb}
\usepackage{multirow}

\usepackage{algorithm} 
\usepackage{listings} 
\usepackage[table]{xcolor} 
\usepackage{pifont} 

\newcommand{\app}{\raise.17ex\hbox{$\scriptstyle\sim$}}
\definecolor{deemphcolor}{gray}{0.6}
\newcommand{\deemph}[1]{\textcolor{deemphcolor}{#1}}

\usepackage[pagebackref=true,breaklinks=true,letterpaper=true,colorlinks,bookmarks=false]{hyperref}

\usepackage[abs]{overpic}
\iccvfinalcopy

\newcommand\blfootnote[1]{%
  \begingroup
  \renewcommand\thefootnote{}\footnote{#1}%
  \addtocounter{footnote}{-1}%
  \endgroup
}

\begin{document}

\title{Contrastive Tuning: A Little Help to Make Masked Autoencoders Forget}

\author{
Johannes Lehner \footnotemark[1]~$~^{1}$\\\and
Benedikt Alkin \footnotemark[1]~$~^{1}$\\
\and
Andreas F\"{u}rst $~^{1}$\\
\and
Elisabeth Rumetshofer $~^{1}$\\
\and
Lukas Miklautz \footnotemark[2]~$~^{2}$$~^{3}$\\
\and
Sepp Hochreiter$~^{1}~^{4}$\\
}

\maketitle

\ificcvfinal\thispagestyle{empty}\fi
\blfootnote{}
\blfootnote{* Equal contribution ~~ \dag~ Main work conducted during research stay}
\blfootnote{$^{1}$~ELLIS Unit Linz and LIT AI Lab, Institute for Machine Learning}
\blfootnote{~~~Johannes Kepler University, Linz, Austria}
\blfootnote{$^{2}$~Faculty of Computer Science, University of Vienna, Vienna, Austria}
\blfootnote{$^{3}$~UniVie Doctoral School Computer Science, University of Vienna}
\blfootnote{$^{4}$~Institute of Advanced Research in  Artificial Intelligence (IARAI)}
\blfootnote{}
\blfootnote{Corresp. to~\{alkin, lehner\}@ml.jku.at, \mbox{lukas.miklautz@unvie.ac.at}}
\blfootnote{Code available at 
 \url{https://github.com/ml-jku/MAE-CT}}

\begin{abstract}
Masked Image Modeling (MIM) methods, like Masked Autoencoders (MAE), efficiently learn a rich representation of the input. However, for adapting to downstream tasks, they require a sufficient amount of labeled data since their rich features code not only objects but also less relevant image background.
In contrast, Instance Discrimination (ID) methods focus on objects. 
In this work, we study how to combine the efficiency and scalability of MIM with the ability of ID to perform downstream classification in the absence of large amounts of labeled data.
To this end, we introduce Masked Autoencoder Contrastive Tuning (MAE\nobreakdash--CT), a sequential approach that utilizes the implicit clustering of the Nearest Neighbor Contrastive Learning (NNCLR) objective to induce abstraction in the topmost layers of a pre-trained MAE.
MAE\nobreakdash--CT tunes the rich features such that they form semantic clusters of objects without using any labels. Notably, MAE\nobreakdash--CT does not rely on hand-crafted augmentations and frequently achieves its best performances while using only minimal augmentations (crop \& flip).
Further, MAE\nobreakdash--CT is compute efficient as it requires at most 10\% overhead compared to MAE pre-training.
Applied to large and huge Vision Transformer (ViT) models, MAE\nobreakdash--CT excels over previous self-supervised methods trained on ImageNet in linear probing, $k$-NN and low-shot classification accuracy as well as in unsupervised clustering accuracy.
With ViT-H/16 MAE\nobreakdash--CT achieves a new state-of-the-art in linear probing of 82.2\%.
\end{abstract}

\begin{figure}[t!]
\begin{center}
\input{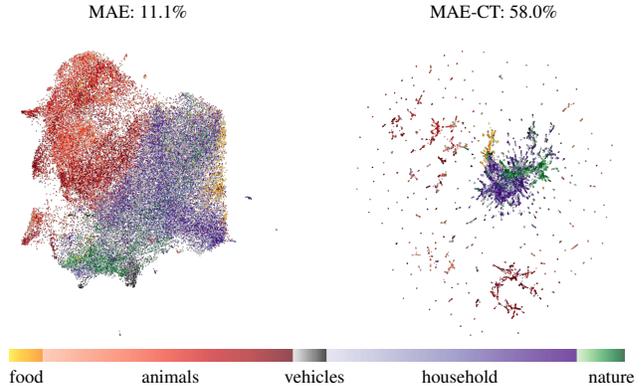}
\end{center}
\caption{Formation of clusters induced by Masked Autoencoder Contrastive Tuning (MAE--CT). MAE leads to a coarse clustering of objects, in the form of ImageNet super-classes, like food, animals, vehicles, household objects, or objects in nature, that summarize several classes. MAE--CT forms object-specific clusters and thereby improves the $k$-means cluster accuracy w.r.t.\ the 1000 ImageNet ground truth classes from 11.1\% to 58.0\% without labeled data or image augmentations based on expert knowledge. 
}
\label{fig:tease}
\end{figure}

\section{Introduction} 

Self-supervised learning (SSL) leverages a pre-training task on unlabeled data to construct rich representations of the input
without explicit supervision from costly annotated labels.
This pre-trained representation can then be used to solve downstream tasks, like image classification, better than supervised training only. Therefore, SSL is currently one of the most effective machine learning concepts. 

Two of the most prominent SSL pre-training tasks in computer vision are \textit{Instance Discrimination} (ID) and \textit{Masked Image Modeling} (MIM). 
ID uses augmentations to create multiple views of an image. The objective is then to align the views created from the same image. To avoid the trivial solution of mapping all images to a constant representation, methods either use a contrastive loss term \cite{he2020momentum,chen2020simple}, a regularization term \cite{zbontar2021barlow} or perform self-distillation \cite{grill2020bootstrap,caron2021emerging}. When a contrastive loss term is used, the ID task can be viewed  as a classification task where each image is its own class.
MIM methods, like Masked Autoencoders (MAE) \cite{he2022masked} and others \cite{bao2021beit,xie2022simmim,  baevski2022data2vec} first mask out areas of the input and then reconstruct the missing parts as pre-train task.

The respective pre-training tasks, classification and reconstruction, of ID and MIM result in distinct advantages and disadvantages. 
MAE provides a computationally efficient way to exploit sparse pre-training of Vision Transformers (ViT)~\cite{dosovitskiy2021image} by masking large parts of the image (75\%) and not processing the masked areas. 
This computational efficiency, coupled with the data efficiency of a generative reconstruction task~\cite{xie2023data, el2021large}, enabled beneficial scaling to larger architectures on datasets of limited size. However, to perform well on downstream tasks, MIM methods rely on fine-tuning with a large amount of labeled data 
as the representation of MIM methods \textit{lack abstraction} 
after pre-training. In contrast, ID methods learn an object-focused representation \cite{caron2021emerging} that typically results in object-specific clusters which is especially useful when few labels are available, as decision boundaries can be drawn much easier 
between well separated clusters.
Furthermore, ID methods notoriously rely on augmentations based on expert knowledge \cite{tian2020makes} to alleviate the problem of \textit{shortcut learning} \cite{geirhos2020shortcut}, which refers to the phenomenon of overfitting to spurious features. 
For example, models could easily learn to extract the color histogram of images to solve the ID task well. To learn more semantic features than a color histogram, color augmentations are essential for almost all ID methods.  
MIM suffers less from this issue as all masked parts have to be reconstructed, leaving little room for shortcut solutions. In fact, MAE achieves its best performance when only minimal augmentations (crop \& flip) are used. 

\begin{figure}[t]
\begin{center}
\includegraphics[width=0.99\linewidth]{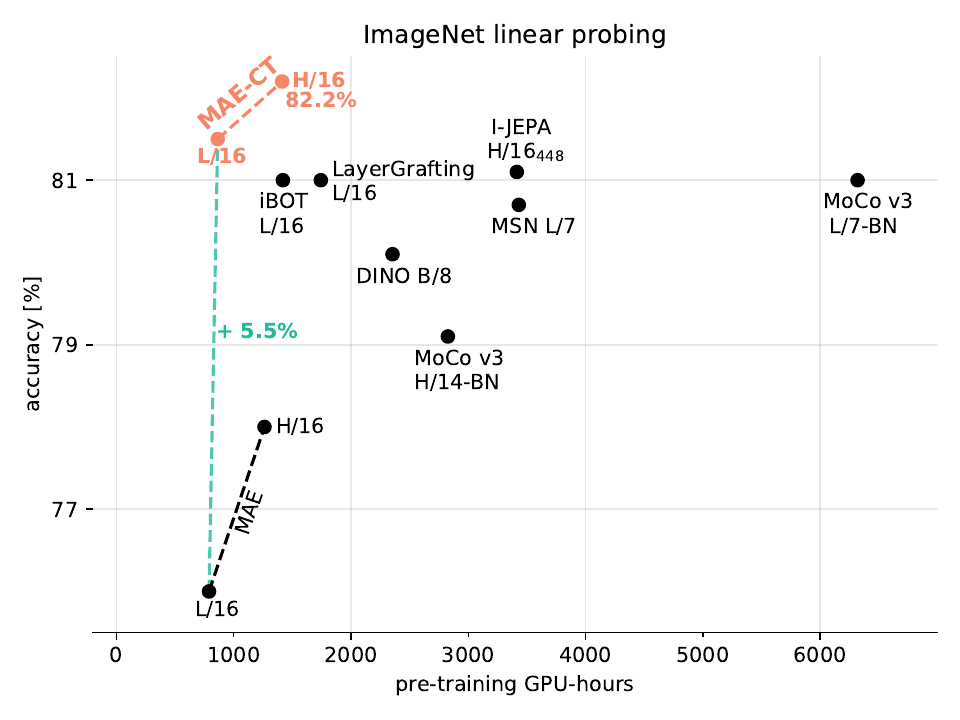}
\end{center}
\caption{ ImageNet linear probing of the best reported models from other self-supervised methods. MAE--CT is able to form a well-clustered representation using little compute.}
\label{fig:gpuhoursvs_probing}
\end{figure} 

\begin{figure}[t]
\begin{center}
\includegraphics[width=0.99\linewidth]{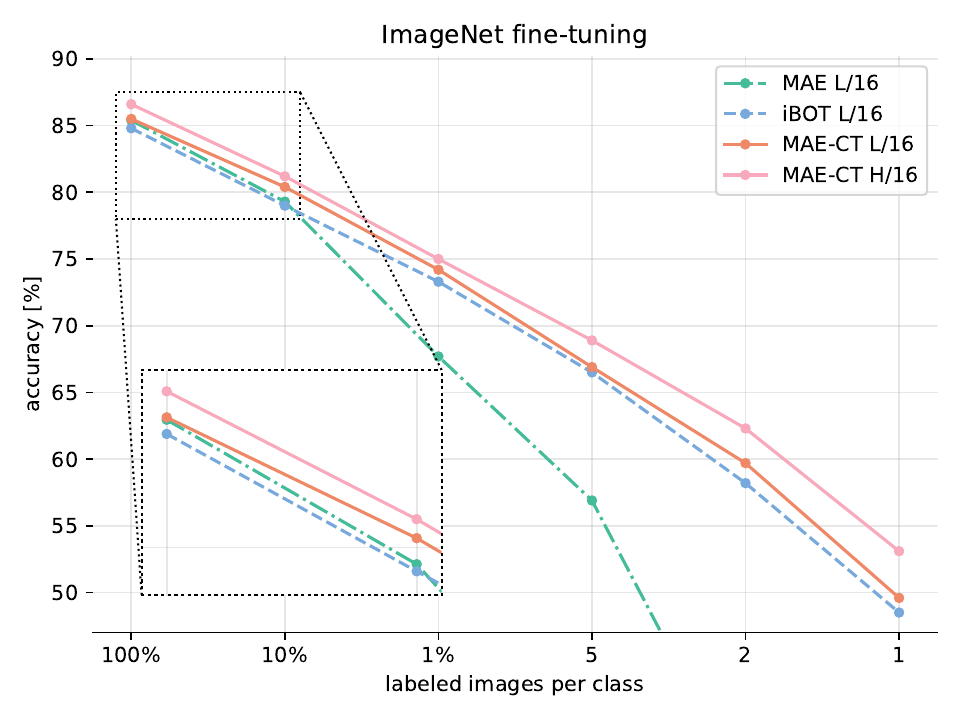}
\end{center}
\caption{ Fine-tuning evaluation on ImageNet. We compare against the best publicly available ID model (iBOT L/16). MAE achieves good performances when given enough labels but struggles otherwise. MAE--CT is able to surpass both MAE and iBOT on all benchmarks.  The improvement increases when considering a similar computational budget (MAE--CT H/16) instead of equal model size. }
\label{fig:finetuning_full_plus_lowshot}
\end{figure}

Given the benefits and downsides of both approaches, an open question remains: What is the best combination of MIM and ID methods to exploit their respective strengths?
Namely, use the unlabeled data and compute efficiency of MIM methods to benefit from larger models while also benefiting from the label efficiency of ID methods for good low-shot performance via a meaningful semantic representation.
Also, extensive augmentations should be optional as they restrict the field of applications and make the models invariant to potentially useful information.  
The straightforward combination of MIM and ID in an end-to-end fashion faces the issue that the objectives and hyperparameters are heavily conflicting. For example, MIM benefits from high masking ratios and minimal augmentations while ID benefits from less masking and extensive augmentations.

We propose a sequential self-supervised approach to combine MIM and ID methods named Masked Autoencoder Contrastive Tuning (MAE\nobreakdash--CT). Contrastive tuning (CT) aims to imitate fine-tuning in the absence of labeled training samples.
MAE\nobreakdash--CT utilizes a contrastive objective to guide a pre-trained MAE encoder to form semantic clusters, visualized in Figure~\ref{fig:tease}.
Unlike previous works, MAE\nobreakdash--CT uses the contrastive objective not to learn basic features, but to induce abstraction in the top-half layers of the pre-trained ViT model. 
This novel setting benefits from multiple adaptations.

State-of-the-art ID methods heavily rely on compute-heavy techniques like multi-crop augmentation~\cite{caron2020swav}, a momentum encoder~\cite{he2020momentum} and long training schedules. We do not rely on these compute-heavy techniques and CT takes only a small fraction of the pre-training duration, consequently, MAE\nobreakdash--CT adds only little overhead to MAE pre-training as depicted in Figure~\ref{fig:gpuhoursvs_probing}.

Nevertheless, we find that the disentanglement of feature learning (via MAE pre-training) and abstraction (via contrastive tuning) produces well separated 
clusters. Thus, MAE\nobreakdash--CT enables a more label efficient downstream classification than the best ID methods, see Figure~\ref{fig:finetuning_full_plus_lowshot}. 

Finally, we adapt the contrastive method Nearest Neighbor Contrastive Learning (NNCLR)~\cite{dwibedi2021}. 
NNCLR extends SimCLR\cite{chen2020simple} with a queue that holds feature vectors of past samples and a Nearest Neighbor (NN) lookup operation. We observe that this feature space augmentation of NNCLR in combination with our sequential approach and scaling of the model size renders the use of extensive input augmentations based on expert knowledge optional rather than mandatory.

\paragraph{We provide the following contributions:}
\begin{enumerate} 

\item We introduce MAE\nobreakdash--CT, a novel computationally efficient and scaleable approach, to form object-related clusters in the representations of pre-trained MAE encoders.

\item We demonstrate that our compute efficient sequential approach is able to surpass the label efficiency of state-of-the-art ID methods in downstream classification.

\item We find that combined pre-training with minimal augmentations suffers from short-cut learning, providing further evidence for the need of our sequential approach. 

\end{enumerate}

\begin{figure*}[ht]
\begin{center}
\begin{overpic}[width=0.95\textwidth]{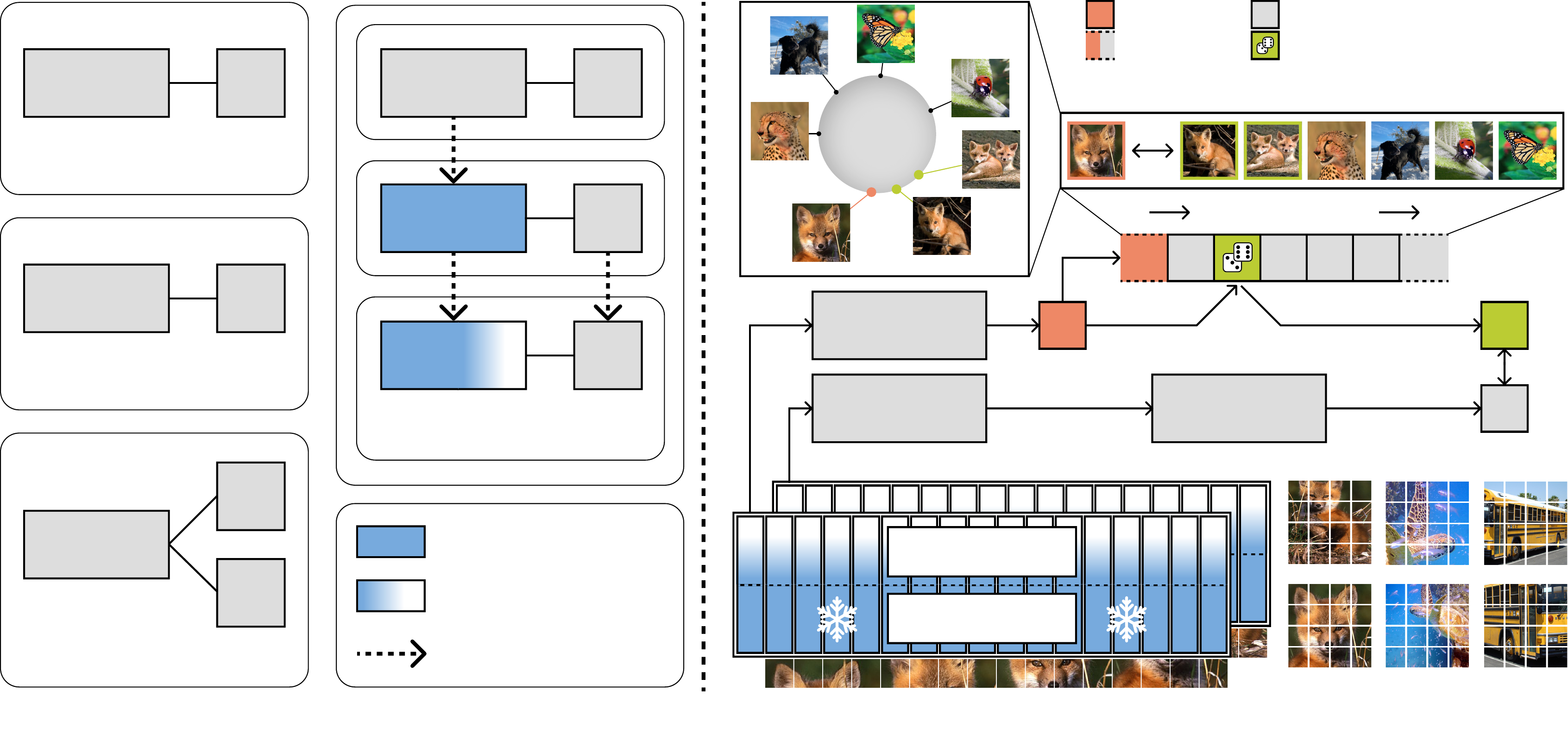}
\begin{scriptsize}
    \put(10,167){\textsf{masked image modeling}}
    \put(16,193.8){\textsf{encoder}}
    \put(69,193.8){\textsf{MIM}}
    \put(10,102){\textsf{instance discrimination}}
    \put(16,128.8){\textsf{encoder}}
    \put(72,128.8){\textsf{ID}}
    \put(30,19){\textsf{combined}}
    \put(16,55.5){\textsf{encoder}}
    \put(69,70.2){\textsf{MIM}}
    \put(72,41.3){\textsf{ID}}
    \put(123,195){\textsf{encoder}}
    \put(176,195){\textsf{MIM}}
    \put(123,154){\textsf{encoder}}
    \put(179,154){\textsf{ID}}
    \put(123,112){\textsf{encoder}}
    \put(179,112){\textsf{ID}}
    \put(127,88){\textsf{\textbf{contrastive tuning}}}
    \put(132,57){\textsf{frozen layers}}
    \put(132,40){\textsf{LR decay}}
    \put(132,23){\textsf{copy weights}}
    \put(28,4){\textsf{\textbf{end-to-end}}}
    \put(124,4){\textsf{\textbf{sequential (MAE-CT)}}}
    
    \put(330,4){\textsf{\textbf{contrastive tuning}}}
    \put(256,96.4){\textsf{projector}}
    \put(256,122){\textsf{projector}}
    \put(359,96.4){\textsf{predictor}}
    \put(432,104){$z_i^+$}
    \put(301,128){$z_i$}
    \put(385,128){$\text{topk-NN} (z_i, Q, k)$}
    \put(461,109){$\mathcal{L}_{i}^\text{NNCLR}$}
    \put(372,156){\textsf{queue} $Q$}
    \put(338,216){\textsf{query}}
    \put(338,205){\textsf{shift in/out}}
    \put(388,216){\textsf{sample in $Q$}}
    \put(388,205){\textsf{selecting sample from topk}}
    \put(475,31){$x_i$}
    \put(475,62){$x_i^+$}
    \put(276,33){\textsf{frozen layers}}
    \put(272,53){\textsf{trainable layers}}
\end{scriptsize}
\end{overpic}
\caption{ Contrary to end-to-end methods, MAE--CT is a sequential approach, as depicted on the left side. 
First, an encoder is pre-trained with a MIM objective (MAE). Afterwards, a NNCLR head is initialized on top of said pre-trained encoder by freezing the encoder and training the NNCLR head until its latent representation is well structured. Finally, contrastive tuning is applied for a short duration.
In contrastive tuning, depicted on the right side, we freeze the bottom half of the ViT-encoder and apply a layer-wise learning rate decay for the top half. Two views of one image are generated and then encoded by the ViT. Both encodings are fed into a projector, followed by either a predictor or a topk-NN lookup resulting in the embeddings for the NNCLR loss.
The queue $Q$ is updated with the new embeddings in a first in -- first out manner after each gradient update step.}
\label{fig:overview}
\end{center}
\end{figure*}
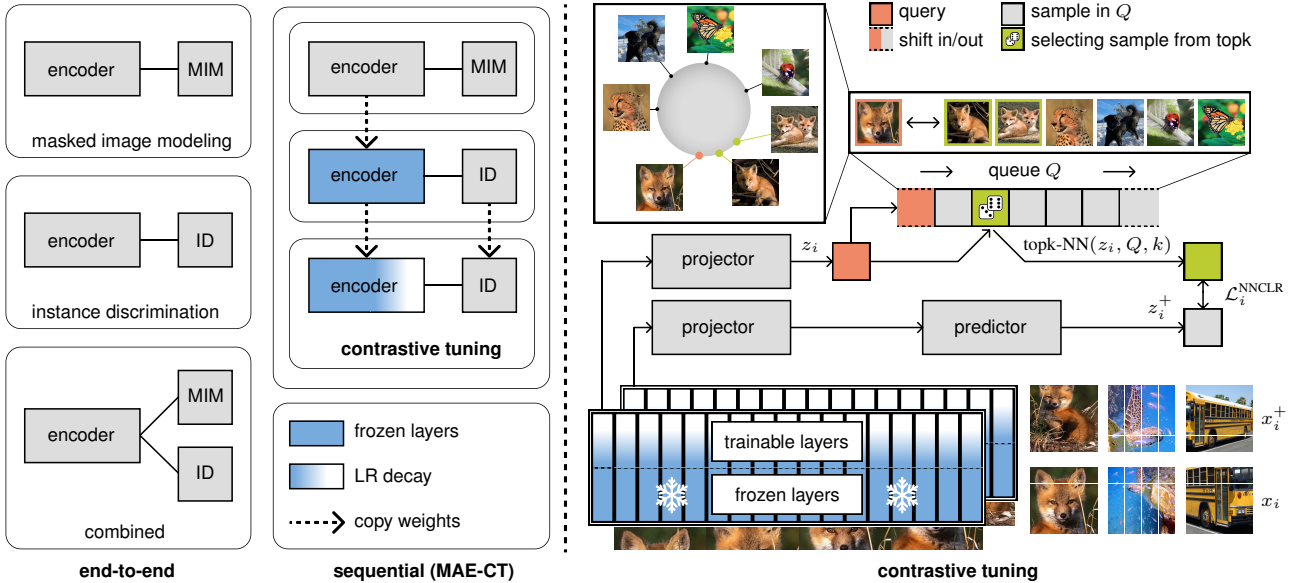

\section{Related Work}
\label{sec:related}
Critical to our sequential approach are the two components MAE~\cite{he2022masked} and NNCLR~\cite{dwibedi2021} on which we build upon. We refer the reader to the respective publications and provide some additional background information in Appendix C. Hereafter, We discuss the most relevant works that make use of both MIM and ID ideas.

The benefit of extending ID with MIM concepts is well established. iBOT~\cite{zhou2021image} adds an auxiliary per-patch reconstruction task. MSN~\cite{assran2022masked} uses masking to improve computational efficiency and augmentation strength. Consequently, these methods report state-of-the-art results in feature and low-shot evaluation. We show that the data efficiency and scaling of MAE enables MAE\nobreakdash--CT to exceed these results.

The advances in MIM motivated multiple works that extend MIM methods with ID concepts. CMAE~\cite{huang2022contrastive} and BootMAE~\cite{dong2022bootstrapped} study the addition of an auxiliary ID objective to MAE pre-training.
Both approaches improve the fine-tuning performance of ViT-B models, 
but at the cost of decreased computational efficiency and increased reliance on expert augmentations. 
They do not report results on larger models, the focus of this work, which would be of interest to measure scalability and data efficiency.

Most related to our approach is the recent work Layer Grafting~\cite{jiang2023layergrafting} which also combines MIM and ID into a sequential approach. MAE pre-training is followed by the full training routine of MoCo~v3~\cite{chen2021mocov3} with an additional regularization loss to keep the lower layer weights close to the MAE weights. 
In contrast to Layer Grafting, MAE\nobreakdash--CT prepares the ID component before changes to the encoder are made, which is then followed up by a short tuning phase. Thus, MAE\nobreakdash--CT is much more compute efficient (see \mbox{Figure~\ref{fig:gpuhoursvs_probing}}). For example, the ID part of Layer Grafting requires 15 times the compute of contrastive tuning with a ViT-L/16.

The goal to pre-train larger ViT models with limited use of hand-crafted augmentations is pursued as well in the recent work I-JEPA~\cite{assran2023self}. I-JEPA models the self-supervised task as a local reconstruction task in feature space. 

\section{Method}
\label{sec:method}

\textit{"An autoencoder wants to remember everything a \mbox{classifier} wants to forget.”} \cite{epstein2019forgetting} 

\paragraph{Motivation.} 
MIM approaches are able to train large ViT models, that lack the inductive bias of convolutional \mbox{neural} networks, and learn rich representations just from ImageNet ~\cite{he2022masked,baevski2022data2vec,xie2022simmim,singh2023effectiveness}. 
However, MIM models rely on adaptation to the downstream tasks using supervised training where  performances heavily degrades as the number of labeled samples decreases. Conversely, ID methods suffer less from this problem as their objective implicitly forms object-specific representations during pre-training\cite{caron2021emerging, walmer2023teaching}. This makes the transition to the downstream task easier as the embedding already represents similar objects in a similar way.
The difference in structure can also be seen by evaluating the embedding directly via linear probing or $k$-NN classification where ID performs significantly better than MIM. Additionally, MAE\nobreakdash--CT is motivated by the reported effectiveness of partial fine-tuning~\cite{he2022masked}. Partial fine-tuning improves classification performance considerably by retraining only a few of the topmost layers in a pre-trained MAE with a supervised objective. This implies that features in the lower layers of a pre-trained MAE already generalize well and that only upper layers need to be tuned. Further, fine-tuning induces an object-specific clustering in the representation of the MAE due to label supervision and achieves invariance to certain input features using a set of extensive input augmentations based on expert domain knowledge.
Finally, fine-tuning adjusts the model from masked inputs used during MAE pre-training to unmasked inputs which are used in downstream tasks. MAE\nobreakdash--CT imitates fine-tuning without labels.

\newpage
\paragraph{Overview.} MAE\nobreakdash--CT is a sequential self-supervised approach to induce abstraction in the representation of a pre-trained MAE. As shown on the left half of Figure~\ref{fig:overview}, MAE\nobreakdash--CT requires three steps.  
First, we perform MAE pre-training. 
Second, we replace the decoder with a NNCLR head.
In an initialization step, the NNCLR head is trained to form fine-grained clusters in the representation that is used for the NN-lookup operation. For this step, the encoder is fully frozen.
Third, the Contrastive Tuning (CT) step, where training effects the weights of the upper half of the encoder and the NNCLR head, depicted on the right side of Figure~\ref{fig:overview} in more detail. During CT the abstract structure of the initialized NNCLR head is transferred back into the encoder to induce a well separated clustering in the encoder output representation. 

\paragraph{MAE pre-training}
MAE pre-training follows the original work~\cite{he2022masked} to learn a rich but coarsely structured representation in a compute efficient manner by randomly masking out a large fraction of the input patches. We do not apply any masking in the subsequent steps.

\paragraph{NNCLR initialization}
Like supervised fine tuning, we want to substantially change the encoder representation within a short tuning duration. Accordingly, we find it essential to learn a good target structure in the NNCLR head before changing the encoder.
This is achieved by freezing the encoder and then training only the small fully-connected NNCLR network.
Notably, we observe that the NNCLR head is able to map the coarse clusters in the frozen MAE encoder to more fine-grained clusters. 
We explain this by the observation that, even with the encoder fully frozen and a lightweight NNCLR head, the contrastive objective is able to form a representation with high uniformity in the NNCLR head despite low uniformity in the encoder representation. 

As discussed in \cite{wang2020hyper}, the contrastive objective can be formulated as the combined minimization of an alignment loss and a uniformity loss. Where alignment is measured as the distance between two views from the same instance and uniformity corresponds to the separability of different instances.
Furthermore, the beneficial effect of mapping to a more uniform representation is reported in \cite{trosten2023hubs}. 

\paragraph{Contrastive tuning}
Contrastive tuning (CT) uses the initialized NNCLR head to retrain the \textit{partially frozen} MAE encoder. 
Although we aim to substantially change the encoder representation within a short duration, change has to happen gradually and has to be restricted such that the learned structure in the initialized NNCLR head is not broken.
We achieve this by employing layer-wise learning rate decay  \cite{clark2020electra} in the encoder.
To make the evolution of the entries written into the NNCLR queue more smooth, we use an exponential moving average (EMA) \cite{laine2016temporal,tarvainen2017mean} for the lightweight projector network only~\cite{pham2022pros}.  

To reduce memory and compute requirements, we mimic partial fine-tuning and freeze the lower half of the encoder.

Furthermore, CT has to take the increase in trainable parameters into account. We observe that while the target structure in the NNCLR representation quickly improves at first, it starts to degrade before the transformation of the encoder representation can be completed. 
To delay this degradation, we increase the difficulty of the alignment task. Instead of using the nearest neighbor of a query vector $z_i$ from the NNCLR queue $Q$ for the NN-lookup, we uniformly sample one of the $k$ nearest neighbors. 
We refer to this adaptation as topk-NN lookup. 

\begin{equation}
\label{eqn:nnclr_topk_lookup}
 \text{topk-NN} (z_i, Q, k) := \underset{\{1,..,k\}}{\mathcal{U}}~\bigg( \underset{q \in Q}{\text{topk}}~
 z_i \cdot q \bigg) 
\end{equation}

Let $z^+$ refer to the predictor path. The positive counterpart $z_i^+$ is attracted to a {topk-NN} of an anchor vector $z_i$, while all other $z^+_j$ within the batch are repulsed. Using the temperature $\tau$, we then obtain the updated loss function.

\begin{equation}
\label{eqn:nnclr_loss}
\mathcal{L}_{i}^\text{NNCLR} = -  \log{
\frac
    {\exp{(\text{topk-NN} (z_i, Q, k)\cdot z_i^+/\tau)}}
    {\sum\limits_{j=1}^n\exp{(\text{topk-NN} (z_i, Q, k)\cdot z_j^+ / \tau)}} 
}
\end{equation}

In contrast to the result of an ablation study in \cite{dwibedi2021}, we find that topk-NN lookup improves performance during CT. We argue that this is enabled by the high quality of the initialized NNCLR latent representation on top of the pre-trained MAE features. Consequently, we can increase the strength of the data-driven augmentation effect from the {topk-NN} lookup by using a higher value for $k$ during CT. 

\section{Experiments and Analysis}
\label{sec:experiments}
\begin{table*}[t]
\begin{center}
\begin{tabular}{ll|ccccc|cc}
& & \multicolumn{5}{c|}{\textit{low-shot evaluations}} & \multicolumn{2}{c}{\textit{feature evaluations}} \\
Architecture & Method & 1 shot & 2 shot & 5 shot & 1\% & 10\% & Linear probing & k-NN \\
\hline
\multirow{6}{*}{ViT-B/16}
& MAE~\cite{he2022masked} & 14.0 & 27.1 & 43.1 & 54.2 & 73.4 & 68.0 & 51.1 \\
& MoCo v3~\cite{chen2021mocov3} & 37.4 & 47.7 & 57.3 & 63.4 & 74.7 & 76.7 & 72.6 \\
& MSN~\cite{assran2022masked} & \textbf{50.3} & \textbf{58.9} & \textbf{65.5} & 69.5 & 75.5 & 77.7  & 76.3\\
& iBOT~\cite{zhou2021image} & 45.3 & 55.5 & 64.3 & \textbf{71.0} & 77.4 & \textbf{79.5} & \textbf{77.1} \\
& Layer Grafting~\cite{jiang2023layergrafting} & 40.0 & 50.2 & 59.3 & 65.5 & \textbf{77.8} & 77.7 & 75.4 \\
& MAE-CT$_{min}$ (ours) & 31.1 & 38.9 & 47.8 & 56.6 & 73.3 & 73.5 & 64.1 \\
& MAE-CT$_{aug}$ (ours) & 37.5 & 47.9 & 57.3 & 63.3 & 74.6 & 76.9 & 73.4 \\
\hline
\multirow{6}{*}{ViT-L/16}
& MAE~\cite{he2022masked} & 14.3 & 34.9 & 56.9 & 67.7 & 79.3 & 76.0 & 60.6 \\
& MSN~\cite{assran2022masked} & 47.5 & 55.5 & 62.5 & 67.0 & 71.4 & 77.3 & 76.2 \\
& iBOT~\cite{zhou2021image} & 48.5 & 58.2 & 66.5 & 73.3 & 79.0 & 81.0 & 78.0 \\
& Layer Grafting~\cite{jiang2023layergrafting} & 47.8 & 57.6 & 65.3 & 69.3 & 80.1 & 81.0 & 77.3 \\
& MAE-CT$_{min}$ (ours) & \textbf{51.8} & \textbf{60.3} & 66.7 & 72.6 & 79.7 & 80.2 & 78.0\\
& MAE-CT$_{aug}$ (ours) & 49.6 & 59.7 & \textbf{66.9} & \textbf{74.2} & \textbf{80.4} & \textbf{81.5} & \textbf{79.1} \\
\hline
\multirow{3}{*}{ViT-H/16}
& MAE~\cite{he2022masked} & 9.0 & 16.4 & 55.2 & 70.0 & 80.8 & 78.0 & 61.1 \\
& MAE-CT$_{min}$ (ours) & \textbf{53.1} & \textbf{62.3} & \textbf{68.9} & \textbf{75.0} & \textbf{81.2} & 81.5 & 79.4 \\
& MAE-CT$_{aug}$ (ours) & 50.1 & 60.2 & 67.7 & \textbf{75.0} & 81.0 & \textbf{82.2} & \textbf{79.8} \\
\end{tabular}
\end{center}
\caption{ Low-shot and feature evaluations for different model sizes on ImageNet. "1 shot" corresponds to 1 label per class. "1\%" is approximately "13 shot". 
MAE-CT$_{aug}$ consistently improves low-shot and feature evaluation performance. As models get larger and fewer labels are used, MAE-CT$_{min}$ outperforms methods that use extensive augmentations with crop \& flip only.
}
\label{tab:lowshot}
\end{table*}

\subsection{Evaluation}
We evaluate our approach via image classification on \hbox{ImageNet}~\cite{deng2009imagenet}, 
where we vary the number of used labels from 100\% down to a single label per class. 
To ensure a fair comparison, we exclude results that are based on additional training data or larger sequence lengths (via higher input resolution or smaller patch size). As a lot of large-scale models are not publicly available, we compare MAE\nobreakdash--CT to the reported results in Appendix~\ref{tab:results_ext}.

We evaluate MAE\nobreakdash--CT when using only minimal image augmentations (MAE\nobreakdash--CT$_{min}$) and when using the same augmentations as in BYOL~\cite{grill2020bootstrap} (MAE\nobreakdash--CT$_{aug}$).

We choose the evaluation protocol based on the number of available labels in accordance to previous works. For evaluating the representation using 100\% of the labels, we train a linear probe and a $k$-NN classifier. With 10\% and 1\% of the labels, we fine-tune the encoder and in the extreme low-shot settings (\textless 1\% labels), we report the accuracy of a logistic regression classifier averaged over three splits. The detailed protocols can be found in Appendix~\ref{sec:appendix_evaluation}.

\subsection{Implementation details}
We outline the most important implementation details and provide all further information in Appendix~\ref{sec:appendix_implementation_details}.

\paragraph{MAE pre-training.} 
We train for $1600$ epochs with a learning rate of $1.5e-4$ and use the "normalize pixels" variant of the MAE loss, which applies a patch-wise normalization to the target pixels before the mean-squared-error loss. 

\paragraph{NNCLR initialization.}
Following~\cite{dwibedi2021}, we use a $3$\nobreakdash-layer MLP as projector, a $2$-layer MLP as predictor and a queue $Q$ of length $65536$.
To initialize the NNCLR-head, we train for 20 epochs on the output of the fully frozen pre-trained MAE encoder with a learning rate of $1e-4$, a temperature $\tau$ of $0.15$ and the default top1-NN lookup. 

\paragraph{Contrastive tuning.}
We use a learning rate of $1e-4$ and apply layer-wise learning rate decay~\cite{clark2020electra} with decay factor $0.65$ to the upper half of the ViT blocks while freezing the lower half. For MAE\nobreakdash--CT$_{min}$, we train ViT-B/L for $20$ epochs and ViT-H for $30$ epochs. For MAE\nobreakdash--CT$_{aug}$ we train ViT-B for $80$ epochs and ViT-L/H for $40$ epochs.

\subsection{Results}

\paragraph{Feature evaluations.} The right column of Table~\ref{tab:lowshot} shows that
MAE\nobreakdash--CT improves the linear separability of MAE features considerably on all model sizes. Even larger gains can be observed when a simple distance based $k$-NN classifier is used.
Although, we find that CT does not lift a MAE pre-trained ViT-B/16 model to the performance level of the best ID methods. 
An increase in model capacity to ViT-L/16 enables MAE\nobreakdash--CT to outperform said ID methods, leveraging the scaleability of MAE pre-training.
With ViT-H/16, our sequential approach even exceeds models that operate on higher image resolutions or smaller patches and achieves state-of-the-art in linear probing (see Figure~\ref{fig:gpuhoursvs_probing}).

\paragraph{Low-shot evaluation.} The middle column of Table~\ref{tab:lowshot} shows classification accuracy when using only a fraction of the labels. Similar to feature evaluations, MAE\nobreakdash--CT shows superior scaling as it outperforms state-of-the-art ID methods on larger models.
While extensive augmentations are superior for smaller models and more labels, they become less effective as model size grows and the number of labels decreases. For ViT\nobreakdash-L/16 models, MAE\nobreakdash--CT$_{min}$ surpasses the performance of methods that use extensive augmentations on the 1 shot and 2 shot benchmark. With a ViT\nobreakdash-H/16, MAE\nobreakdash--CT$_{min}$ is able to surpass the performance of MAE\nobreakdash--CT$_{aug}$ on all low-shot benchmarks.

\paragraph{Clustering analysis.} We assess the ability of MAE\nobreakdash--CT to form object-specific clusters in two ways. First, we use the cluster accuracy \cite{YangXNYZ10_cluster_accuracy,xie2016unsupervised} to measure how well the ground truth classes of the validation set of ImageNet can be discovered using the unsupervised $k$-means clustering algorithm. The cluster accuracy ranges between 0 and 100, where 100 indicates a perfect match with the ground truth. Second, we calculate the silhouette score \cite{rousseeuw1987silhouettes} to quantify the spread and compactness of the ground truth classes. The silhouette score ranges from -100 to 100, with 100 being the best value. Silhouette scores smaller than zero indicate that the clusters are not well separated. 

Both cluster accuracy and silhouette score are reported in Table~\ref{tab:clustering_metrics}. Compared to MAE, MAE\nobreakdash--CT shows a large improvement in cluster performance. 
The silhouette score improves from being negative, finding almost no cluster structure, to being positive showing separated clusters (see also Figure \ref{fig:tease}). 
On ViT-L/16, MAE\nobreakdash--CT outperforms all other ID methods even when using only minimal augmentations.
To the best of our knowledge, the cluster accuracy of MAE\nobreakdash--CT (ViT-H/16) is state-of-the-art on ImageNet, when trained on ImageNet only. Further details are provided in Appendix~\ref{app:cluster_analysis}.

\begin{table}[ht]
\begin{center}
\begin{tabular}{lccc}
Method & {B/16} & {L/16} & {H/16}\\
\hline
MAE & 13.8 (-5.4) & 14.3 (-4.1) & 11.1 (-7.6)\\
MoCo~v3 & 43.0 (4.5) & - & - \\
MSN & \textbf{54.2} (10.4) & 45.4 (4.8) & - \\
iBOT & 50.0 (6.7) & 52.0 (9.0) & - \\
MAE-CT$_{min}$ & 35.3 (1.1) & 54.9 (11.0) & \textbf{58.0} (8.4) \\
MAE-CT$_{aug}$ & 46.2 (4.3) & \textbf{56.9} (10.1) & 54.8 (7.9) \\
\end{tabular}
\end{center}
\caption{ $k$-means cluster accuracy on ImageNet. Parentheses show the silhouette score w.r.t. the ground truth. }
\label{tab:clustering_metrics}
\end{table}

\subsection{Ablations and analysis}

\paragraph{Ablation.} 
We evaluate the impact of essential CT components in Table~\ref{tab:ablation}. 
Masking during contrastive tuning results in a moderate drop in performance, but enables training on a single GPU.
The NNCLR head initialization, by training it on top of frozen encoder features before CT, is essential to the performance gains. As MAE\nobreakdash--CT does not use any augmentations besides crop and flip, the data-driven augmentation of the NN lookup is required to make the NNCLR task difficult enough to learn a useful representation.

\begin{table}[h]
\begin{center}
\begin{tabular}{lcc}
Method & Probing & $k$-NN \\ 
\hline
MAE\nobreakdash--CT$_{min}$ & \textbf{80.2} & \textbf{77.4} \\ 
apply masking (75\%) during CT & 79.4 & 75.3 \\
skip NNCLR head initialization & 78.5 & 68.1 \\
NNCLR without NN lookup & 70.6 & 40.5 \\ 
\end{tabular}
\end{center}
\caption{ Ablating study of CT components with ViT-L/16. }
\label{tab:ablation}
\end{table}

\newpage
\paragraph{Combined pre-training of MAE and NNCLR.}
In addition to the sequential MAE\nobreakdash--CT approach, we also investigate the combined pre-training of MAE and NNCLR as depicted in the bottom left of Figure~\ref{fig:overview}.
To this end, we train a MAE with a ViT-L/16 encoder by additionally attaching a NNCLR head onto the [CLS] token of the encoder. During training we jointly optimize the MAE and NNCLR objectives where we balance the losses \mbox{$\mathcal{L}$ = $\mathcal{L}^\text{MAE}$ + $\lambda$  $\mathcal{L}^\text{NNCLR}$} 
with a $\lambda$ of 0.001. Note, that we keep the augmentations from MAE, which are crop \& flip only.

The upper half of Table~\ref{tab:combined} shows that the combined pre-training slightly improves over MAE but is far worse than MAE\nobreakdash--CT. 
In addition to just combined pre-training, we investigate the application of CT after combined pre-training. We investigate different ways to initialize the NNCLR head before CT in the lower part of Table~\ref{tab:combined}. For "combined pre-training + CT" we initialize a new NNCLR head by training it on top of frozen encoder features just like we do for MAE\nobreakdash--CT. For "combined pre-training + CT$_\text{skip}$" we reuse the NNCLR head from combined pre-training directly for CT, which effectively \textit{skips} the reinitialization of the NNCLR head. 
Additionally, instead of initializing an NNCLR head on top of the frozen encoder, one can train an NNCLR head during MAE training by inserting a stop gradient operation before the NNCLR head ("detached pre-training + CT$_\text{skip}$"). 
The results show that directly using the NNCLR head from pre-training for CT is slightly worse in addition to being more constrained as it requires modification of the MAE training process.

Overall the sequential approach of MAE\nobreakdash--CT is more flexible, more compute efficient and achieves superior performances compared to the combined pre-training, even when CT is applied in addition to combined pre-training.

\begin{table}[t]
\begin{center}
\begin{tabular}{lcc}
Method & Probing & $k$-NN \\
\hline
MAE & 76.0 & 60.6 \\
combined pre-training & 77.8 & 66.1 \\
\hline
combined pre-training + CT$_\text{skip}$ & 79.7 & 75.3\\ 
detached pre-training + CT$_\text{skip}$ & 80.1 & 77.0 \\  
combined pre-training + CT & 80.1  & 76.7\\ 
MAE\nobreakdash--CT$_{min}$ & \textbf{80.2} & \textbf{77.4} \\  
\end{tabular}
\end{center}
\caption{ Comparison of combined pre-training of MAE and NNCLR without and with applying CT.
}
\label{tab:combined}
\end{table}

\begin{figure}[h]
\begin{center}
\includegraphics[width=0.95\linewidth]{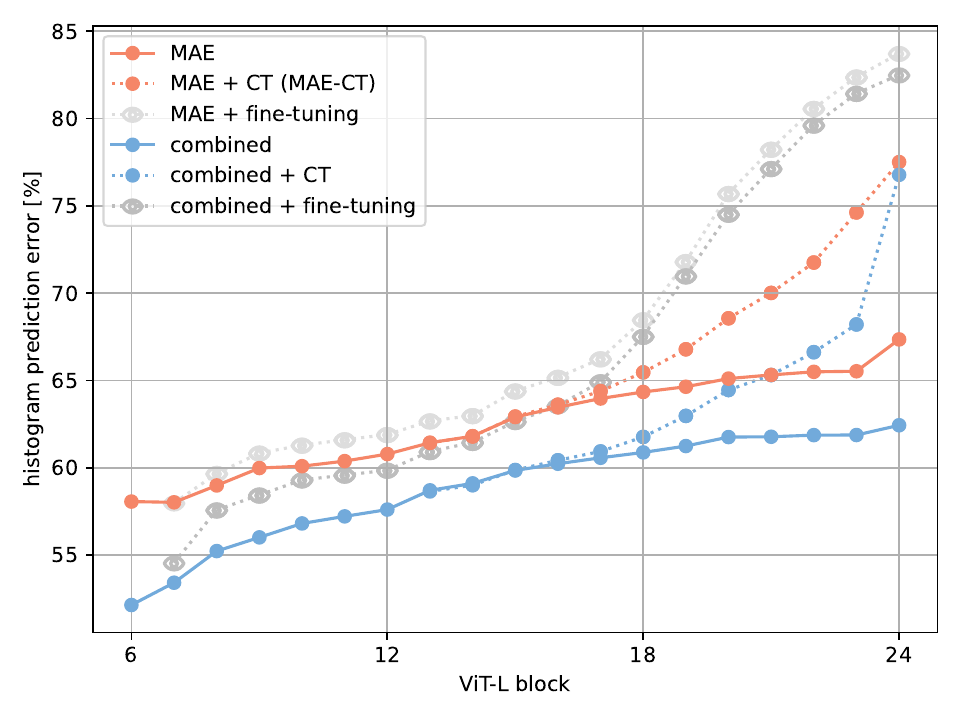}
\end{center}
\caption{Measuring shortcut learning: Predicting the color histogram of input images from the [CLS] token after different blocks of a ViT-L (higher error is better).
Combined pre-training learns shortcut features that can only partially be \textit{forgotten} via CT or supervised fine-tuning. Sequential training avoids shortcut learning.
}
\label{fig:color_hist}
\end{figure}

\paragraph{Shortcut learning of combined pre-training.}
In experiments with combined pre-training, we observe that even with a small  NNCLR loss weight $\lambda$ during combined pre-training, the NNCLR loss decreases immediately by about 30\% compared to training a detached NNCLR head during combined pre-training. As the encoder does not have a good feature representation at the start of training, this indicates that the NNCLR head steers the encoder to extract basic features which drastically simplify the NNCLR objective. These basic features might be a symptom of shortcut learning~\cite{geirhos2020shortcut}.
For ID methods, a form of shortcut learning is to learn color statistics of the input image, as two views of the same image likely have a similar color histogram.
Following \cite{addepalli2022towards} we utilize a prediction task to estimate to what degree the model preserves information about color statistics within the [CLS] token. We train a linear probe to predict the color histograms of the input image. In the combined pre-training, the NNCLR head pushes the encoder towards learning color statistic features (Figure~\ref{fig:color_hist}). These shortcut features evolve already in early encoder layers of the combined pre-training. While CT is able to partially correct them, they remain more dominant than in the sequentially trained encoder. 
Not even supervised fine-tuning of all layers --- using augmentations based on expert knowledge --- can fully mitigate this effect.
We describe the color histogram prediction task in Appendix~\ref{sec:appendix_analysis}.

\begin{table}[h]
\begin{center}
\begin{tabular}{lc|cccc}
 Method & $k$ & V2 & R & Sketch & A  \\
\hline
MAE & - & 71.7 & 37.0 & 24.3  & 12.3 \\ 
MAE\nobreakdash--CT$_{min}$ &  1  & 76.4 & 41.2 & 27.6 & 20.5 \\ 
MAE\nobreakdash--CT$_{min}$ & 10 & \underline{76.6} & 42.0  & 28.2 & 20.8 \\ 
MAE\nobreakdash--CT$_{min}$ & 20 & 76.5 & \underline{42.2} & 28.1 & 21.4  \\ 
MAE\nobreakdash--CT$_{min}$ & 30 & 76.5 & \underline{42.2}  & \underline{28.4} & \underline{21.5} \\ 
MAE\nobreakdash--CT$_{aug}$ & 10 & \textbf{77.9} & \textbf{48.8}  & \textbf{35.8} & \textbf{28.2} \\ 
\end{tabular}
\end{center}

\caption{ Robustness evaluation with linear classifiers trained on ImageNet. We compare different ViT-L/16 models and report accuracy for multiple values of 
 $k$ in the topk-NN lookup.}
\label{tab:transfer_large}
\end{table}

\newpage

\paragraph{Robustness and transfer.} 
We evaluate the linear probe trained on ImageNet on four transfer learning datasets.
Namely, ImageNet-V2~\cite{Recht2019iamgenetv2}, ImageNet-R~\cite{hendrycks2021many}, ImageNet-Sketch~\cite{wang2019learning} and ImageNet-A~\cite{hendrycks2021nae}.
In Table~\ref{tab:transfer_large} we compare the top-1 accuracy of ViT-L/16 models. 
The results show a small, but overall consistent benefit of the stronger augmentation effect provided by higher values of $k$ in the topk-NN lookup for MAE\nobreakdash--CT. MAE\nobreakdash--CT$_{aug}$ is more robust to distribution shifts. The gap between MAE\nobreakdash--CT and MAE\nobreakdash--CT$_{aug}$ becomes larger with increase in distribution shift of the datasets (from left to right), but always improves MAE.

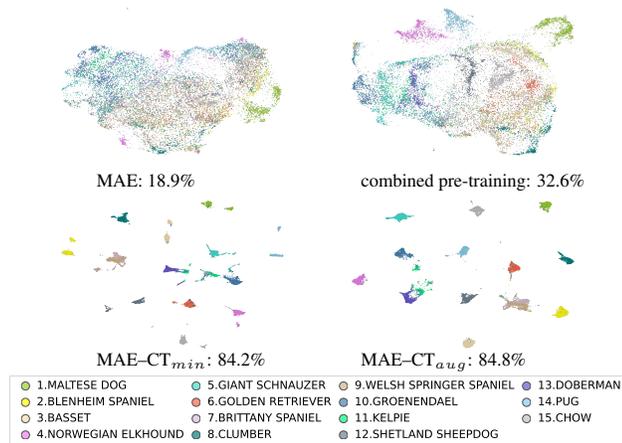
\begin{figure}[b]
\begin{center}
\begin{overpic}[width=\linewidth]{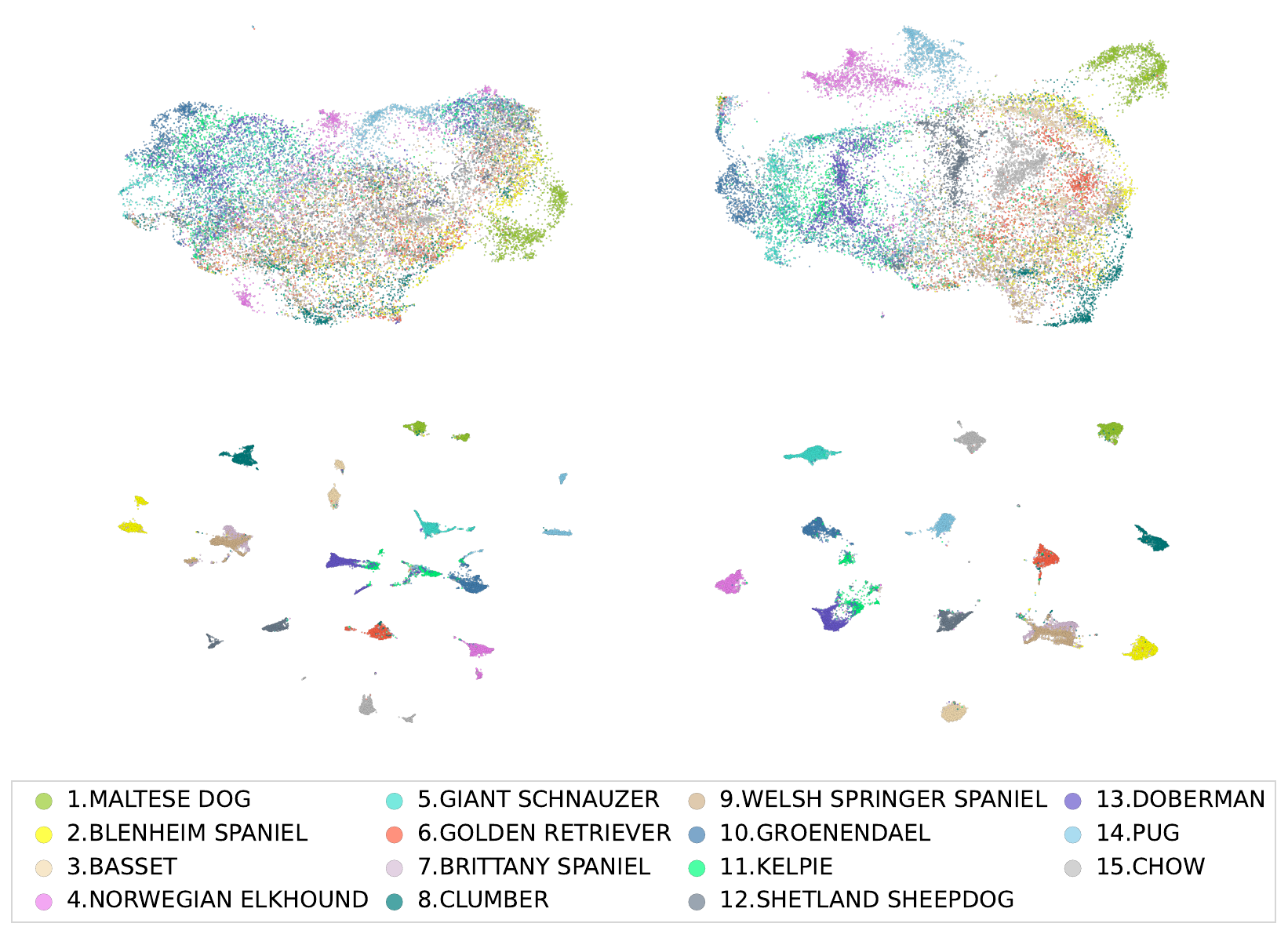}
\begin{scriptsize}
\put(35,100){MAE: 18.9\%}
\put(135,100){combined pre-training: 32.6\%}
\put(35,32){MAE\nobreakdash--CT$_{min}$: 84.2\%}
\put(135,32){MAE\nobreakdash--CT$_{aug}$: 84.8\%}
\end{scriptsize}
\end{overpic}
\end{center}
\caption{ UMAP embeddings of MAE, combined pre-training, MAE\nobreakdash--CT$_{min}$ and MAE\nobreakdash--CT$_{aug}$ with corresponding $k$-means cluster accuracies for ImageNet-Dogs15 (ViT-L). MAE\nobreakdash--CT clearly improves the separation of the 15 classes. }
\label{fig:dogs_umap}
\end{figure}

\paragraph{Cluster formation.} 
To demonstrate the differences in clustering we provide results for ImageNet-Dogs15~\cite{ChangWMXP17_imagenetdogs}, a subset of ImageNet commonly utilized in the clustering literature. In Figure \ref{fig:dogs_umap} we show the UMAP~\cite{mcinnes2018umap-software} embedding of different variations (ViT\nobreakdash-L) with their corresponding clustering accuracy. 
Combined pre-training finds well separated clusters for the classes \textit{Norwegian Elkhound}, \textit{Pug} and \textit{Maltese Dog}.
We suspect that these three dog breeds have only small intra-class variations in their characteristics, which makes them easily discernible by low level features. Once contrastive tuning is applied the cluster accuracy improves by a factor of four and also the classes are visually better separated. 

\begin{figure}[h]
\begin{center}
\includegraphics[width=\linewidth]{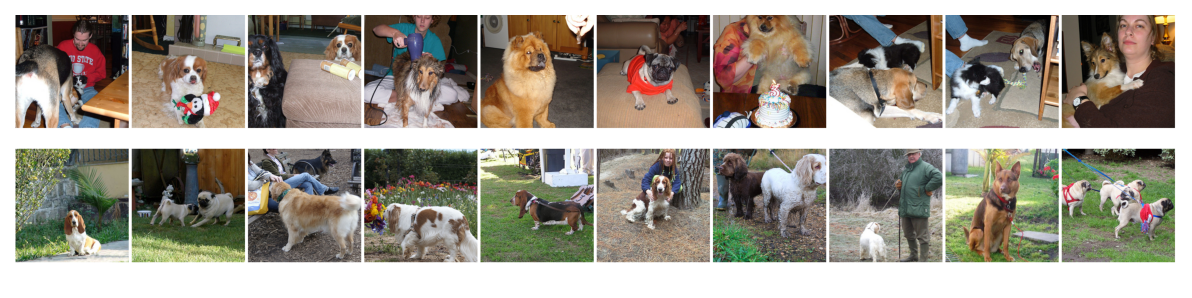} \\
 \includegraphics[width=\linewidth]{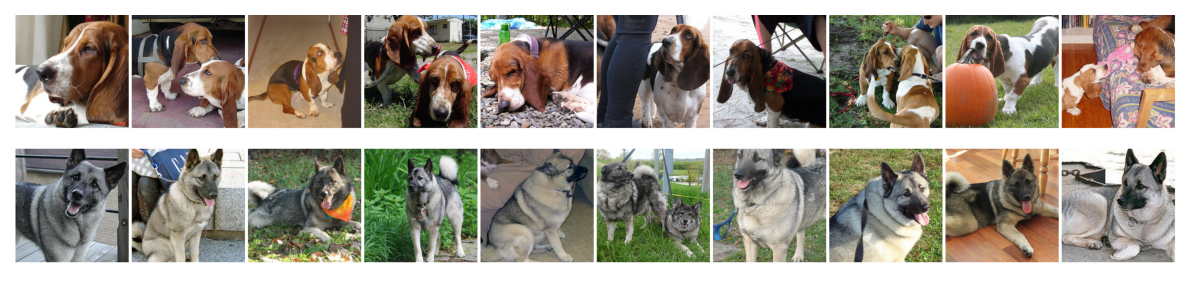}
\end{center}
   \caption{Ten NNs for two $k$-means cluster centers for MAE (upper) and  MAE\nobreakdash--CT$_{min}$ (lower). Each row corresponds to one cluster found in the [CLS] token of ViT-H/16 for ImageNet-Dogs15. MAE groups the images into dogs located indoors (first row) and outdoors (second row) depending on the background. MAE\nobreakdash--CT finds clusters that correspond to the specific dog breeds. 
   }
\label{fig:cluster_retrievals}
\end{figure}

\paragraph{Cluster retrieval.} 
Figure \ref{fig:cluster_retrievals} shows the NNs of two $k$\nobreakdash-means cluster centroids for ImageNet-Dogs15 of MAE and  MAE\nobreakdash--CT$_{min}$. Inspecting the NNs of the cluster centroid indicates that MAE finds some clusters that correspond to image backgrounds, the first row contains dogs that are located inside and the second row contains dogs that are outside. This is also quantified by the low cluster accuracy w.r.t. the ground truth dog breeds of MAE. MAE reaches a cluster accuracy of 18.7\% vs. 94.3\% reached by MAE\nobreakdash--CT. MAE\nobreakdash--CT finds distinct clusters containing mostly images of a single class, shown by the perfect NN retrievals of the classes \textit{Basset} (third row) and \textit{Norwegian Elkhound} (fourth row). Note, that the dogs are correctly grouped despite the different background.
We provide the full retrieval, confusion matrices and UMAP embeddings in Appendix~E. 

\section{Conclusion}
\label{sec:conclusion}
We introduce MAE\nobreakdash--CT, a self-supervised approach to combine the strengths of MIM and ID methods. 
We show that the NNCLR training objective --- applied to an already pre-trained MAE model --- is capable of creating object-specific clusters in its feature representation which greatly improves representation quality (linear probing, $k$-NN and cluster accuracy) and low-shot classification performance. 

We show that our sequential approach preserves the data \mbox{efficiency} of MAE and incorporates the label efficiency of NNCLR while requiring only 10\% more compute than MAE pre-training. This allows us to train large ViT models on ImageNet only where our larger models exceed the performance of previous state-of-the-art SSL models.

In contrast to state-of-the-art ID methods, MAE\nobreakdash--CT does not rely on hand-crafted image augmentation. 
This is a very promising result, which can be explained by the data-driven augmentation effect of the NN-lookup, which greatly benefits from representations that already capture image semantics in a structured way.

\clearpage
\section*{Acknowledgements}
The ELLIS Unit Linz, the LIT AI Lab, the Institute for Machine Learning, are supported by the Federal State Upper Austria. IARAI is supported by Here Technologies. We thank the projects AI-MOTION (LIT-2018-6-YOU-212), DeepFlood (LIT-2019-8-YOU-213), Medical Cognitive Computing Center (MC3), INCONTROL-RL (FFG-881064), PRIMAL (FFG-873979), S3AI (FFG-872172), DL for GranularFlow (FFG-871302), EPILEPSIA (FFG-892171), AIRI FG 9-N (FWF-36284, FWF-36235), ELISE (H2020-ICT-2019-3 ID: 951847), Stars4Waters (HORIZON-CL6-2021-CLIMATE-01-01). We thank Audi.JKU Deep Learning Center, TGW LOGISTICS GROUP GMBH, Silicon Austria Labs (SAL), FILL Gesellschaft mbH, Anyline GmbH, Google, ZF Friedrichshafen AG, Robert Bosch GmbH, UCB Biopharma SRL, Merck Healthcare KGaA, Verbund AG, GLS (Univ. Waterloo) Software Competence Center Hagenberg GmbH, T\"{U}V Austria, Frauscher Sensonic and the NVIDIA Corporation

We thank the European High Performance Computing initiative for providing  the computational resources that enabled this work. 
EHPC-DEV-2022D09-012, 
EHPC-DEV-2022D09-024, 
EHPC-DEV-2022D06-166, 
EHPC-DEV-2022D06-177,
EHPC-DEV-2022D06-173 

{\small
\bibliographystyle{ieee_fullname}
\bibliography{arxiv2}
}
\clearpage
\appendix
\section*{Appendix}
To ensure reproducibility, we list all implementation details regarding training (A) and evaluation (B). We discuss relevant background information (C). We present additional experiments and results (D). Finally, we provide a cluster analysis and show additional qualitative results (E).

\section{Training Details}
\subsection{Implementation details}
\label{sec:appendix_implementation_details}
 \paragraph{General} All experiments use the linear $\textit{lr}$ scaling rule~\cite{LinearLRScalingRule-Warmup} $\textit{lr}=\textit{base lr} \times \textit{batch size} \times \textit{views}/256$ where $\textit{views}=2$ for CT and $\textit{views}=1$ otherwise. 
Bias and normalization parameters (weights of \mbox{LayerNorm} and BatchNorm layers) are excluded from the weight decay.

In early experiments with float16 we frequently encountered overflows in the attention matrix leading to $\inf$ activations. Therefore, we train all our models in bfloat16~\cite{Kalamkar2019bfloat} precision.
We use FlashAttention~\cite{FlashAttention}, which speeds up training by approximately 40/30/20\% for ViT-B/L/H respectively.

\paragraph{MAE reimplementation.} In early experiments, we investigated combined pre-training of MAE and NNCLR with a stop gradient operation between encoder and NNCLR head. This effectively corresponds to normal MAE training as the NNCLR head \textit{does not} influence the MAE pre-training. The only difference is that two views (generated with the augmentations used for MAE pre-training) are used. To account for this difference, we reduce the number of pre-training epochs from 1600 to 800. Additionally, we train with a lower batch size due to memory constraints ($1024$ for ViT-B/L and $512$ for ViT-H). We train a ViT-B/16, ViT-L/16 and ViT-H/16 using this procedure, which we use throughout the main paper. For ViT-H/14 we use the publicly available checkpoint. Table~\ref{tab:mae_reimpl} shows that the performance of our re-implementation is nearly identical to the original.

\begin{table}[h]
\begin{center}
\begin{tabular}{lcc}
Method & Probing & k-NN \\ 
\hline
MAE$_{public}$ & 75.8 & 60.6 \\
MAE$_{reimpl}$ & 75.9 & 57.6 \\
\hline
MAE$_{public}$-CT$_{min}$ & 80.2 & 77.3 \\
MAE$_{reimpl}$-CT$_{min}$ & 80.2 & 77.4 \\
\end{tabular}
\end{center}
\caption{ Validation of our MAE re-implementation by comparing to the original publicly released checkpoint of a ViT-L/16. For this experiment, CT is done with $k=1$ in the topk-NN lookup.}
\label{tab:mae_reimpl}
\end{table}

\newpage

\begin{algorithm}[h]
   \caption{PyTorch-style pseudo-code description of \hbox{contrastive} tuning using an adapted NNCLR head}
   \label{alg:contrastive_tuning}
    \definecolor{codeblue}{rgb}{0.25,0.5,0.5}
    \lstset{
      basicstyle=\fontsize{7.2pt}{7.2pt}\ttfamily\bfseries,
      commentstyle=\fontsize{7.2pt}{7.2pt}\color{codeblue},
      keywordstyle=\fontsize{7.2pt}{7.2pt},
    }
\begin{lstlisting}[language=python]
# f: ViT encoder .. lower half of the encoder is frozen
# g: projector .. 3 layer MLP, hidden 2048, output 256
# h: predictor .. 2 layer MLP, hidden 4096, output 256 
# Q: queue .. shift register of length 65536
# f_m: slow EMA of encoder f .. the model used after CT
# t1: 0.9999 .. momentum parameter of f_m
# g_m: fast EMA of projector g  
# t2: 0.99 .. momentum parameter of g_m

# load a minibatch x 
for x in loader:  
    # two randomly augmented versions of x
    x1, x2 = augment(x), augment(x)
    # encoder forward passes
    y1, y2 = f(x1), f(x2)
    # predictor path for each y
    p1, p2 = h(g(y1)), h(g(y2))
    # NN path for each y
    z1, z2 = g_m(y1), g_m(y2)
    # L2-normalize embeddings
    p1, p2 = normalize(p1, dim=1), normalize(p2, dim=1)
    z1, z2 = normalize(z1, dim=1), normalize(z2, dim=1)
    # topk-NN lookup for each z
    nn1 = topk_NN(z1, Q, k) 
    nn2 = topk_NN(z2, Q, k) 
    # symmetrised InfoNCE loss 
    loss = L(nn1, p2)/2 + L(nn2, p1)/2
    # optimization step
    loss.backward()
    optimizer.step()
    # shift embeddings z1 into queue Q
    update_queue(Q, z1)
    # EMA weight updates
    f_m = t1 * f_m + (1 - t1) * f
    g_m = t2 * g_m + (1 - t2) * g

# topk-NN lookup
def topk_NN(z, Q, k):
     similarities = z @ Q.T # matrix multiplication 
     # calculate list of topk-NN
     candidates = similarities.topk(k, dim=1).indices
     # uniform sampling
     dice = torch.randint(size=(length(Q),), high=k) 
     idx = candidates[torch.arange(length(Q)),dice]
     return Q[idx]

# InfoNCE loss
def L(nn, p, temperature=0.15):
    logits = nn @ p.T # matrix multiplication
    logits /= temperature # sharpening
    labels = torch.arange(p.shape[0]) 
    loss = cross_entropy(logits, labels)
    return loss

\end{lstlisting}
\end{algorithm}

\subsection{Contrastive tuning}
\label{sec:appendix_contrastive_tuning}

\paragraph{ViT architecture.}
We follow MAE~\cite{he2022masked} and use the standard ViT architecture~\cite{dosovitskiy2021image}. The ViT projects the input images into patch tokens with a non-overlapping convolutional layer ($\text{stride}=\text{kernel size}$), adds a fixed 2D sine-cosine positional embedding~\cite{vaswani2017transformer} to each patch token, appends an auxiliary [CLS] token and then processes all tokens with a stack of transformer blocks. 
A transformer block consists of two modules, namely a multi-head self-attention and a 2-layer MLP, which are applied sequentially. Both modules are preceded by a LayerNorm~\cite{ba2016layernorm} layer and are wrapped with a residual connection~\cite{he2016resnet}. After the last transformer block, a final LayerNorm is applied. We use the [CLS] token of the encoder output as input to the NNCLR head and all evaluations. For ViT-B/16, we use the average of the patch tokens instead of the [CLS] token.

\paragraph{NNCLR architecture.}
We follow the design of the original NNCLR head~\cite{dwibedi2021} and use a 3-layer MLP as projector and a 2-layer MLP as predictor. Each MLP layer is followed by a BatchNorm~\cite{BatchNorm} except the last predictor layer. Each BatchNorm layer is followed by a ReLU activation except the last projector layer. BatchNorm statistics are calculated per view. 
To \mbox{stabilize} the entries written into the queue, we apply a fast moving exponential moving average (EMA) \cite{he2020momentum, tarvainen2017mean} to the projector MLP in the topk-NN lookup path. 

\paragraph{NNCLR initalization}
Before contrastive tuning, we train an NNCLR head with the encoder fully frozen to adjust the NNCLR head to the encoder features using top1-NN lookup, a learning rate of $1e-4$, no EMA in the projector, a temperature $\tau$ of 0.15 and a training duration of $20$ epochs for all model sizes and variants. Other hyperparameters follow the ones from contrastive tuning (Table~\ref{tab:hyperparams_tuning} and~\ref{tab:hyperparams_aug_tuning}).

\paragraph{Contrastive tuning.}
Algorithm~\ref{alg:contrastive_tuning} describes the process of contrastive tuning in a PyTorch-like pseudo code.
Table~\ref{tab:hyperparams_tuning} and \ref{tab:hyperparams_aug_tuning} list the hyperparameters used during contrative tuning of MAE\nobreakdash--CT$_{min}$ and MAE\nobreakdash--CT$_{aug}$, respectively. 
We find higher values for the temperature $\tau$ to perform best with ViT-H models. We argue that this is caused by the increased capacity of ViT-H and that a higher $\tau$ controls the alignment and uniformity trade-off~\cite{wang2020hyper} in the InfoNCE loss. Therefore, a higher $\tau$ slows down the expansion of clusters, as observed in~\cite{wang21understanding}.
It is common practice in student-teacher ID methods to use the teacher version of the encoder after training \cite{caron2021emerging}. 
Thus, we use an exponential moving average (EMA) version of the encoder after contrastive tuning.

\begin{table}
\begin{center}
\begin{tabular}{l|ccc}
MAE\nobreakdash--CT &  Base & Large & Huge  \\
\hline
Epochs & 20 & 20 & 30 \\ 
Batch size & 1024 & 1024 & 512 \\
Optimizer & \multicolumn{3}{c}{AdamW~\cite{loshchilov2018adamw}
} \\
\quad Momentum & \multicolumn{3}{c}{$\beta_1=0.9,\beta_2=0.95$} \\
Learning rate schedule & \multicolumn{3}{c}{warmup $\rightarrow$ cosine} \\
\quad Warmup fraction~\cite{szegedy2016rethinkinginception}
& 20 \% & 20 \% & 20 \% \\
Encoder &  &  &  \\ 
\quad Learning rate & 1e-4 & 1e-4 & 1e-4 \\ 
\quad Layer-wise lr decay~\cite{clark2020electra}
& 0.65 & 0.65 & 0.65 \\
\quad Weight decay & 0.05 & 0.05 & 0.05\\
\quad Frozen layers & 6 & 12 & 16 \\ 
\quad EMA & 0.9999 & 0.9999 & 0.9999 \\
NNCLR head & \\
\quad Learning rate & 1e-4 & 1e-4 & 1e-4 \\ 
\quad Weight decay & 1e-5 & 1e-5  & 1e-5 \\
\quad Temperature $\tau$ & 0.15 & 0.2 & 0.3  \\ 
\quad topk-NN $k$ & 20 & 20 & 20 \\
\quad Projector EMA & 0.99 & 0.99 & 0.995 \\

\end{tabular}
\end{center}
\caption{ MAE\nobreakdash--CT$_{min}$ hyperparameters.}
\label{tab:hyperparams_tuning}
\end{table}

\begin{table}
\begin{center}
\begin{tabular}{l|ccc}
MAE\nobreakdash--CT$_{aug}$ & Base & Large & Huge  \\
\hline
Epochs & 80 & 40 & 40 \\ 
Batch size & 1024 & 1024 & 512 \\
Optimizer & \multicolumn{3}{c}{AdamW~\cite{loshchilov2018adamw}
} \\
\quad Momentum & \multicolumn{3}{c}{$\beta_1=0.9,\beta_2=0.95$} \\
Learning rate schedule & \multicolumn{3}{c}{warmup $\rightarrow$ cosine} \\
\quad Warmup fraction~\cite{szegedy2016rethinkinginception}
& 20 \% & 20 \% & 20 \% \\
Encoder & & & \\
\quad Learning rate & 1e-4 & 1e-4 & 1e-4 \\ 
\quad Layer-wise lr decay~\cite{clark2020electra}
& 0.65 & 0.65 & 0.65 \\
\quad Weight decay & 0.05 & 0.05 & 0.05\\
\quad Frozen layers & 6 & 12 & 16 \\ 
\quad EMA & 0.9999 & 0.9999 & 0.9999 \\
NNCLR head & & & \\
\quad Learning rate & 5e-4 & 5e-4 & 5e-4 \\
\quad Weight decay & 1e-5 & 1e-5  & 1e-5 \\
\quad Temperature $\tau$ & 0.15 & 0.15 & 0.35  \\ 
\quad topk-NN $k$ & 10 & 10 & 30 \\
\quad Projector EMA & 0.99 & 0.99 & 0.995 \\
\end{tabular}
\end{center}
\caption{ MAE\nobreakdash--CT$_{aug}$ hyperparameters.}
\label{tab:hyperparams_aug_tuning}
\end{table}

\section{Evaluation Details}
\label{sec:appendix_evaluation}

\paragraph{GPU hours benchmark.}
\label{benchmark_details}
As a comparison between runtimes is often difficult due to differences in code, software and hardware setups we re-implement the compare methods and measure the runtime via a short benchmark run. Benchmarks are conducted on a single A100 40GB PCIe card using pytorch 2.0 with CUDA 11.8 using bfloat16~\cite{Kalamkar2019bfloat} precision and FlashAttention~\cite{FlashAttention}. We use the largest possible batchsize that is a power of 2. The total number of GPU-hours in Figure~2 is then extrapolated to the number of epochs reported in the original works. As there exists no official implementation for the ViT BatchNorm variants of MoCo~v3, we instead report the runtime of the corresponding LayerNorm variant. 

\paragraph{Linear probing.}
We follow common linear probing protocols and sweep over learning rates~\cite{zhou2021image,caron2021emerging,assran2023self} and insert a non-affine BatchNorm~\cite{BatchNorm} layer after the fully frozen encoder~\cite{doersch2015unsupervised,he2022masked,assran2022masked}. Adding a BatchNorm layer does \textit{not} break the linearity property as it can be absorbed into the linear layer after training. Table~\ref{tab:linear_probing_hyperparameters} lists all parameters.

\begin{table}[H]
\begin{center}
\begin{tabular}{l|c}
Config & Value \\
\hline
Epochs & 50 \\
Batch size & 1024 \\
Optimizer & SGD \\
\quad Momentum & 0.9 \\
\quad Weight decay & 0 \\
Learning rate &  \\
\quad Peak & $\{0.1, 0.09, \dots, 0.01\}$ \\
\quad Schedule & warmup $\rightarrow$ cosine \\
\quad Warmup epochs~\cite{LinearLRScalingRule-Warmup}
& 5 \\
Augmentations & \\
\quad \tt{RandomResizedCrop} & $224^2$ \\

\quad \tt{RandomHorizontalFlip} & $p=0.5$ \\
\end{tabular}
\end{center}
\caption{ Linear probing  hyperparameters. }
\label{tab:linear_probing_hyperparameters}
\end{table}

\paragraph{$k$-NN classification.}
We also evaluate the feature representation via a weighted $k$-NN classifier following the protocol in DINO~\cite{wu2018unsupervised,caron2021emerging}. 
We sweep over different $k$ for every compared method and find $k=10$ to perform well for MAE\nobreakdash--CT and MAE\nobreakdash--CT$_{aug}$. For public checkpoints that do not report $k$-NN accuracies in their original publication, we report the best accuracy over different $k$ for each checkpoint (which is also $k=10$ most of the time).
The weighted $k$\nobreakdash-NN classifier takes the $k$ training samples with the highest cosine similarity and weights each neighbors class by its cosine similarity with the test sample.

\paragraph{Low-shot classification.}
When using 10\% of the ImageNet labels, we use the low-shot fine-tuning protocol of I-JEPA~\cite{assran2023self} to evaluate all compared ID methods (Table~\ref{tab:lowshot_finetuning_hyperparameters}). 
We find that all MAE models, all MAE\nobreakdash--CT models and the huge models of MAE\nobreakdash--CT$_{aug}$ benefit from more regularization in this setting and therefore use the protocol from the original MAE publication~\cite{he2022masked}, which uses Mixup~\cite{zhang2018mixup}, Cutmix~\cite{yun2019cutmix}, RandomErase~\cite{zhong2020RandomErasing}, DropPath~\cite{huang2016stochasticdepth} (0.1/0.2/0.3 for ViT-B/L/H) in addition to \mbox{RandAugment}~\cite{nips2020randaug} as well as a higher learning rate ($1e-3$). We also consider the MAE fine-tuning protocol for ID methods but find the I-JEPA protocol to perform better. 

In the 1\% label regime, the protocols of related work vary between fine-tuning~\cite{assran2023self} and logistic regression~\cite{caron2021emerging}. Therefore, we explore logistic regression, the fine-tuning protocol used above and said protocol but with only crop \& flip as input augmentations. We find MAE, MAE\nobreakdash--CT and MAE\nobreakdash--CT$_{aug}$ to consistently perform better with only crop \& flip and report the performance of the best protocol for competing methods. The best performing protocols are: fine-tuning with only crop \& flip for MAE, logistic regression for MoCo v3~\cite{chen2021mocov3}, fine-tuning for iBOT~\cite{zhou2021image} and logistic regression for MSN~\cite{assran2022masked}. Following~\cite{cai2022semivit} we use a higher lr ($1e-2$) for MAE with ViT-H/14.

For low-shot evaluations with 1, 2 or 5 labeled samples per class we train a $\text{L}_2$-regularized logistic regression using the \texttt{cyanure} package~\cite{cyanure} with hyperparameters from previous works~\cite{caron2021emerging,assran2022masked}. As MAE also benefits from fine-tuning in this low-data regime, we additionally explore full fine-tuning and fine-tuning the last encoder block where we find full fine-tuning to perform best overall. In extreme cases (1 and 2 labels per class with ViT-H) fine-tuning the last block and logistic regression perform slightly better, but we report the full fine-tuning results for consistency.

\begin{table}[ht]
\begin{center}
\begin{tabular}{l|c}
Config & Value \\
\hline
Epochs & 100 (B), 50 (L/H) \\
Batch size & 512 \\
Optimizer & AdamW \\
\quad Momentum & $\beta_1=0.9, \beta_2=0.999$ \\
\quad Weight decay & 0.05 \\
Learning rate &  \\
\quad Peak & 3e-5 \\
\quad Schedule & warmup $\rightarrow$ cosine \\
\quad Warmup epochs~\cite{LinearLRScalingRule-Warmup}
& 5 \\
Layer-wise lr decay~\cite{clark2020electra}
& 0.65 (B), 0.75 (L/H) \\
DropPath~\cite{huang2016stochasticdepth}
& 0.0 \\
Label smoothing~\cite{szegedy2016rethinkinginception}
& 0.1 \\
Augmentations & \\
\quad \tt{RandomResizedCrop} & $224^2$\\
\quad \tt{RandomHorizontalFlip} & $p=0.5$ \\
\quad \tt{RandAugment}~\cite{nips2020randaug}
& \\
\qquad \tt{num\_layers} & 2 \\
\qquad \tt{magnitude} & 9 \\
\end{tabular}
\end{center}
\caption{ Low-shot fine-tuning hyperparameters. }
\label{tab:lowshot_finetuning_hyperparameters}
\end{table}

\subsection{Analysis}
\label{sec:appendix_analysis}
\paragraph{Color histogram prediction.}
We evaluate the amount of information about color statistics by constructing a color histogram prediction task. We calculate the color histograms of an input image, discretize it into $64$ bins and normalize it such that the sum of all bins of each color channel is equal to $1$. This results in a total of $192$ values which we regress with a L1 loss. To make the loss interpretable, we normalize it by dividing it by the average loss on the ImageNet validation set when predicting a uniform distribution (which evaluates to  approximately $0.01204$). We call the average over these $192$ normalized loss values "histogram prediction error", which can be interpreted as a percentage where $0\%$ is achieved by perfectly predicting the histograms and $100\%$ is the random performance.

We concurrently train $24$ linear probes using the [CLS] tokens of each of the $24$ ViT-L layers with a slightly modified version of the setting described in Table~\ref{tab:linear_probing_hyperparameters}. We train for $10$ epochs with $1$ warmup epoch and a learning rate of $0.1$. The encoder is fully frozen throughout training.

\section{Background}

\label{sec:background_appendix}
\paragraph{Instance discrimination (ID).} 
First, ID methods generate different views of the same image via augmentations. 
Then, they align the representations of these views in order to obtain representations that are invariant to the transformations that generated the views~\cite{tian2020makes}. 
A trivial solution for alignment would be a constant representation, which corresponds to 
a undesired collapse of the representation space to a single point.
Representation collapses can be avoided in different ways: 
(a) contrastive methods \cite{hjelmlearning,he2020momentum,chen2020simple,misra2020self} 
rely on a loss that penalizes the similarity to views of other samples, the so-called negatives; 
(b) teacher-student approaches \cite{grill2020bootstrap, caron2021emerging, chen2021exploring} 
prevent collapses via an asymmetry between an active student and a teacher; 
(c) various methods circumvent collapses by constraints that rely on clustering \cite{caron2018deep, caron2020swav} or on invariances \cite{zbontar2021barlow, bardesvicreg}.
Typically, ID methods learn object-specific representations that are suited to solve downstream 
classification tasks with few labels.

\paragraph{Masked Image Modeling (MIM).} 
MIM uses a pre-training task where missing or corrupted parts of the input image have to be reconstructed.

Neither early image in-painting methods~\cite{doersch2015unsupervised, pathak2016context} nor the adaptation of masked language modeling~\cite{devlin2019bert} in computer vision~\cite{chen2020generative} achieved the performance of supervised pre-training. 
In computer vision, masking methods were revived by the introduction of the 
Vision Transformer (ViT; \cite{dosovitskiy2021image}).
Due to their missing inductive biases, ViTs struggle to learn local features with limited data from supervised training alone~\cite{raghu2021dovision}. 
Furthermore, in contrast to convolutional neural networks, the ViT architecture readily enables to mask out patches without introducing undesired artifacts. 

Various methods have shown that MIM can improve data efficiency and allows scaling of the model size \cite{bao2021beit, he2022masked,xie2022simmim,  baevski2022data2vec}. 
On smaller datasets, enhancing contrastive methods by MIM
was beneficial \cite{el2021large}. 
Furthermore, masking the input of the student network has been used to improve the performance of self-distillation methods \cite{zhou2021image, assran2022masked}.

Masked self-supervised learning is a very general concept and, therefore, can be applied to 
various domains such as natural language processing~\cite{devlin2019bert}, audio processing~\cite{huang2022AudioMAE,PatchOut}, or computer vision~\cite{he2022masked,bao2021beit,chen2020generative,xie2022simmim}.
However, MIM pre-training often does not create features that are suited for downstream tasks such as classification via linear probing or $k$-NN. 
Typically, downstream tasks require object-focused features while MIM constructs features that are more general and allow to capture the image background. 
For example, MAEs must also represent the image background to achieve a low reconstruction error.
Therefore, adaptation mechanisms such as end-to-end fine-tuning must be applied to achieve optimal performance on downstream tasks. 
These adaption mechanisms rely on hand-crafted data augmentations~\cite{he2022masked} such as \mbox{Rand}Augment~\cite{nips2020randaug}, Mixup~\cite{zhang2018mixup}, Cutmix~\cite{yun2019cutmix}, 
and other regularization techniques~\cite{huang2016stochasticdepth,clark2020electra}.

\paragraph{Contrastive and clustering methods.}     
Clustering-based contrastive 
methods \cite{li2021contrastive_clustering,caron2018deep,caron2020swav,caron2021emerging} 
leverage group information in the form of clusters or prototypes to avoid splitting objects that belong to the same semantic class. 
Therefore, these methods create object-focused features, which leads to clusters of
objects of the same type or even clusters of different views of the same object.
However, clustering is prone to early over-generalization by considering an entire cluster 
as positive examples before a proper discrimination of samples is possible \cite{dwibedi2021}. 
This phenomenon is known in the deep clustering literature \cite{zhou2022comprehensive}, 
in which a two step approach was shown to avoid over-generalization and 
trivial solutions \cite{caron2018deep,xie2016unsupervised,ji2019invariant}. 
First, pre-training via a self-supervised method learns a suitable representation. 
Subsequently, samples are clustered starting from the learned representation, 
thereby  the model is fine-tuned to induce better separation.
The two step approach can also be implemented
as a joint \cite{xie2016unsupervised} or an alternating \cite{yang2017towards} procedure.

Similar to our approach, many deep clustering methods use 
autoencoders for pre-training \cite{zhou2022comprehensive}, since they
are more generally applicable than contrastive methods. 
However, their reconstruction objective generates features that do not fit to the discriminative nature of 
clustering or classification tasks 
which need object-focused features \cite{epstein2019forgetting,MiklautzBMTBP21} . 

To explain the success of contrastive learning, some work considers the connection between clustering and contrastive methods \cite{parulekar2023infonce,arora2019,haochen2022theoretical,ChenFNZ0FR22}, with recent work \cite{assran2022clusterprior} pointing to a connection to the $k$-means algorithm \cite{macqueen1965_kmeans,lloyd1982_kmeans}.

\paragraph{Masked Autoencoder (MAE).}MAE~\cite{he2022masked} is a popular MIM method, 
as it enables the training of large ViT models on small datasets with reduced computational cost. 
During pre-training, a larger ViT model serves as an encoder and 
a smaller ViT model as a decoder of a Denoising Autoencoder~\cite{vincent2008denoising}. 
The latter regresses the missing patches. 
The high masking ratio shortens the sequence length processed in the encoder, which vastly reduces the computational cost of pre-training. 
MAEs exhibit strong performance when fine-tuned on 
computer vison downstream tasks, such as object detection or instance segmentation~\cite{li2021benchmarking,hu2022LongSequenceMAE}, and also on other modalities such as  audio~\cite{huang2022AudioMAE} or video~\cite{feichtenhofer2022masked, tongvideomae}. 

\paragraph{Nearest Neighbor Contrastive Learning (NNCLR).}
NNCLR \cite{dwibedi2021} extends the popular contrastive method 
\mbox{SimCLR}~\cite{chen2020simple} with a Nearest Neighbor (NN) lookup. 
Instead of calculating the InfoNCE loss between an anchor sample of one view and a positive sample of the other view, the feature vector of the positive sample is replaced by its nearest neighbor from a queue containing feature vectors of previous samples. The queue is implemented via a shift register (similar to MoCo~\cite{he2020momentum}). 
Additionally, NNCLR adopts the asymmetric head design of BYOL~\cite{grill2020bootstrap}.

When restricted to only crop \& flip augmentations, NNCLR shows promising resilience to shortcut learning~\cite{geirhos2020shortcut}.
In an experiment with reduced training duration, NNCLR reports a relative drop of only 2.1\% in linear probing performance with ResNet50~\cite{he2016resnet}.
For comparison, other ID methods report a relative drop of 
\app 13\%~(BYOL~\cite{grill2020bootstrap}), 
\app 27\% (Barlow Twins~\cite{zbontar2021barlow}), 
or \app 28\%~(SimCLR~\cite{chen2020simple}) in performance. \\

\section{Additional Results \& Experiments}

\label{sec:sensitivity_study}
\paragraph{Hyperparameter sensitivity.} We evaluate the sensitivity of chosen hyperparameters on a ViT-L/16 in Table~\ref{tab:sensitivity}. MAE\nobreakdash--CT works well across a range of hyperparameters.

\begin{table}[h]
\begin{center}
\begin{tabular}{lcccc}
{\text{Training epochs}}  &{10} & \cellcolor{gray!15}{20} & {30} & -  \\
{\text{Probing accuracy}} & 79.57 & \cellcolor{gray!15}\textbf{80.23} & \textbf{80.24}  & - \\
{\text{$k$-NN accuracy}} & 76.23 & \cellcolor{gray!15}\textbf{77.36} & 76.81  & - \\
\hline
{\text{Trainable layers}} & {6} & \cellcolor{gray!15}{12} & {15}   & {18} \\
{\text{Probing accuracy}} &  {79.76} & \cellcolor{gray!15}\textbf{80.23} & \textbf{80.26} & {80.12} \\
{\text{$k$-NN accuracy}}& 76.91 & \cellcolor{gray!15}{77.36}  & {77.43} & \textbf{77.49} \\
\hline
\text{Encoder EMA} & {0.0} & {0.999} & \cellcolor{gray!15}{0.9999}  & -  \\
{\text{Probing accuracy}} & {80.19} & \textbf{80.21} & \cellcolor{gray!15}\textbf{80.23}   & - \\
{\text{$k$-NN accuracy}} & {77.29} & {77.23} & \cellcolor{gray!15}\textbf{77.36}   & - \\
\hline
\text{Temperature}~$\tau$ & {0.15} & \cellcolor{gray!15}{0.2} & {0.25}  & - \\
{\text{Probing accuracy}}  & {80.12} & \cellcolor{gray!15}\textbf{80.23} &  {80.12}  & - \\
{\text{$k$-NN accuracy}} & {77.18}  & \cellcolor{gray!15}\textbf{77.36} & \textbf{77.39}  & - \\
\hline
\text{$k$-NN lookup}~$k$ & \cellcolor{gray!15}{1} & {10} & {20} & {30} \\
{\text{Probing accuracy}} & \cellcolor{gray!15}{80.23} & {80.16} & {80.17} &  \textbf{80.29} \\
{\text{$k$-NN accuracy}} & \cellcolor{gray!15}{77.36} & {77.98} & \textbf{77.98} &  {77.90} \\
\hline
\text{Projector EMA} & {0.0} & \cellcolor{gray!15}{0.99} & {0.995} & {0.999} \\
{\text{Probing accuracy}} & {80.07} & \cellcolor{gray!15}\textbf{80.23} & {80.16} & {80.13} \\
{\text{$k$-NN accuracy}} & {77.13} & \cellcolor{gray!15}\textbf{77.36} & {77.31} & {77.08} \\
\end{tabular}
\end{center}
\caption{ Sensitivity analysis of MAE\nobreakdash--CT with ViT-L/16 using only crop \& flip augmentations. Default parameters used in this analysis are marked in \colorbox{gray!15}{gray}.}
\label{tab:sensitivity}
\end{table}

\paragraph{Representations for linear probing.} Following previous works~\cite{caron2021emerging,devlin2019bert},
we evaluate the representation of the concatenation of the features from the last $l$ ViT blocks as well as the concatenation of the [CLS] token with the average pooled patch tokens. Table~\ref{tab:concat_probe} shows  that $l=6$ works well for our larger models while the concatenation of the [CLS] and the average of the patch tokens works best for ViT-B/16.
We provide these results solely for the sake of completeness and to provide a point of reference. We do not apply this protocol when we compare to other methods.

\begin{table}[h]
\begin{center}
\begin{tabular}{llcccc|c}
\multicolumn{2}{l}{concat $l$ last layers} & 1 & 2 & 4 & 6 & 1$^{\dag}$ \\
\hline
MAE\nobreakdash--CT$_{min}$ & B/16 & 73.5 & 73.2 & 74.2 & 74 & \textbf{74.5} \\
MAE\nobreakdash--CT$_{min}$ & L/16 & 80.2 & 80.3 & 80.5 & \textbf{80.8} & 80.3 \\
MAE\nobreakdash--CT$_{min}$ & H/16 & 81.5 & \textbf{81.6} & \textbf{81.6} & \textbf{81.6} & \textbf{81.6} \\
\hline
MAE\nobreakdash--CT$_{aug}$ & B/16 & 76.9 & 75.5 & 76.0 & 75.9 & \textbf{77.5} \\
MAE\nobreakdash--CT$_{aug}$ & L/16 & 81.5 & 81.6 & \textbf{81.8} & \textbf{81.8} & 81.6 \\
MAE\nobreakdash--CT$_{aug}$ & H/16 & 82.2 & 82.2 & \textbf{82.4} & 82.3 & 82.3 \\
\end{tabular}
\end{center}
\caption{ ImageNet linear probing performance when concatenating the features of the last $l$ layers. $^{\dag}$: concatenation of the [CLS] token and the average of all patch tokens of the last layer.} 
\label{tab:concat_probe}
\end{table}

\paragraph{Fine-tuning with 100\% of the labels.} 
We show the results for our models in Table~\ref{tab:finetune_100p}. As the margin between MAE and MAE\nobreakdash--CT is small, we report mean and standard deviation of three seeds. Note that all MAE\nobreakdash--CT models start from our own MAE models (as described in Section~\ref{sec:appendix_implementation_details}) which perform slightly worse in the 100\% labels benchmark than the original models.

\begin{table}[h]
\begin{center}
\begin{tabular}{lcccc}
Method & ViT-B/16 & ViT-L/16 & ViT-H/16 \\ 
\hline
\deemph{MAE$_{public}$} & \deemph{83.57} & \deemph{85.87 (0.07)} & \deemph{-} \\
MAE$_{reimpl}$ & 83.20 & 85.37 (0.05) & 86.50 (0.01) \\
MAE-CT$_{min}$ & 83.32 & 85.47 (0.03) & 86.59 (0.05) \\
MAE-CT$_{aug}$ & 83.22 & 85.56 (0.10) & 86.57 (0.08) \\
\end{tabular}
\end{center}
\caption{ Fine-tuning with 100\% of the labels. Parenthesis show the mean and standard deviation of three seeds. }
\label{tab:finetune_100p}
\end{table}

\begin{table*}
\begin{center}
\begin{tabular}{ll|ccccc|cc}
& & \multicolumn{5}{c|}{\textit{low-shot evaluations}} & \multicolumn{2}{c}{\textit{feature evaluations}} \\
Architecture & Method & 1 shot & 2 shot & 5 shot & 1\% & 10\% & Linear probing & k-NN \\
\hline
\multirow{3}{*}{ViT-H/16}
& MAE & 9.0 & 16.4 & 55.2 & 70.0 & 80.8 & 78.0 & 61.1 \\
& MAE-CT$_{min}$ & \textbf{53.1} & \textbf{62.3} & \textbf{68.9} & \textbf{75.0} & \textbf{81.2} & 81.5 & 79.4 \\
& MAE-CT$_{aug}$ & 50.1 & 60.2 & 67.7 & \textbf{75.0} & 81.0 & \textbf{82.2} & \textbf{79.8} \\
\hline
\multirow{4}{*}{ViT-H/14}
& MAE &  7.2 & 14.1 & 40.2 & 72.8 & 81.2 & 77.2 & 58.9 \\
& I-JEPA & 34.1 & 46.3 & 58.1 & 73.7 & 79.5 & 79.3 & 71.6 \\
& MAE-CT$_{min}$ & \textbf{49.4} & \textbf{59.6} & \textbf{67.4} & \textbf{74.4} & 81.3 &  81.2 & \textbf{79.1} \\
& MAE-CT$_{aug}$ & 45.5 & 56.2 & 64.9 & 74.0 & \textbf{81.7} & \textbf{82.0} & 78.9 \\ 
\end{tabular}
\end{center}
\caption{ Low-shot and feature evaluations of ViT-H models on ImageNet. The coupling of patch size with mask size of MAE degrades representation quality as analyzed by~\cite{hu2022LongSequenceMAE}. Given the inferior starting point of MAE H/14, MAE-CT is still able to leverage the longer sequence length when sufficiently many labels are available (10\%), but falls short to the ViT-H/16 model otherwise. We also compare against I-JEPA~\cite{assran2023self}, a recent joint-embedding self-supervised approach. }
\label{tab:lowshot_huge}
\end{table*} 
\paragraph{Results with ViT-H/14.}
As analyzed in LongSequencMAE~\cite{hu2022LongSequenceMAE}, MAE couples the patch size and the mask size since the mask is sampled on a per patch basis. A smaller patch size results in an easier reconstruction task as the average distance between visible patches is reduced. In experiments with ViT-H/14, we find that this coupling leads to a significant degrade in initial representation quality, which manifests in a 2.1\% drop in k-NN accuracy and a 0.8\% drop in linear probing accuracy (Table~\ref{tab:lowshot_huge}). Since MAE\nobreakdash--CT uses the encoder of a pre-trained MAE, it relies on a good encoder representation. While contrastive tuning can achieve similarly large performance gains on ViT-H/14 when compared to MAE, it only surpasses the ViT-H/16 results on the 10\% low-shot benchmark where the number of labels is sufficient to benefit from the increased sequence length.
Despite this special interaction in the MAE pre-training, MAE\nobreakdash--CT is still able to surpass competing methods using the ViT-H/14 model.

\paragraph{Effective invariance to data augmentation.} 
We evaluate the average effective invariance \cite{deng2022} against rotations and different color augmentations.
For an image $x$ and its augmented version $x_t$, 
\cite{deng2022} defines the effective invariance (EI) as $0$ if the linear classifier predicts different classes 
and $\sqrt{\hat{p}_t \cdot \hat{p}}$ if the predictions agree. 
$\hat{p}$ denote the predictions for $x$ and $\hat{p}_t$ denote the predictions for $x_t$. 
Table~\ref{tab:effective_invariance} shows the average EI for rotations against 90°, 180° and 270° and for color augmentations that include changes in brightness, contrast, hue and saturation. We report the average EI for images in the validation set of ImageNet.

CT increases EI regarding rotations, where using extensive augmentations performs similar to not using them.
The hand-crafted image augmentations used in MAE\nobreakdash--CT$_{aug}$ further increase the EI against color augmentations, but also MAE\nobreakdash--CT improves over MAE. These results indicate that MAE\nobreakdash--CT develops increased robustness against color transformations even though during training the color information in the input is not perturbed. 

\begin{table}[h]
\begin{center}
\begin{tabular}{lcccc}
Method          & B/16          & L/16          &   H/16          \\
\hline
\multicolumn{4}{l}{\textit{Effective invariance against rotations}} \\
MAE             & 0.288             & 0.413             & 0.450             \\
MAE\nobreakdash--CT$_{min}$          & \underline{0.433} & \textbf{0.508}    & \textbf{0.530}    \\
MAE\nobreakdash--CT$_{aug}$ & \textbf{0.437}    & \underline{0.505} & \underline{0.529}             \\
\hline
\multicolumn{4}{l}{\textit{Effective invariance against color augmentations}} \\
MAE             & 0.481             & 0.670             & 0.681             \\
MAE\nobreakdash--CT$_{min}$          & \underline{0.544} & \underline{0.719} & \underline{0.730} \\
MAE\nobreakdash--CT$_{aug}$  & \textbf{0.714}    & \textbf{0.783}    & \textbf{0.786}    \\      
\end{tabular}
\end{center}
\caption{
Effective Invariance (EI) against multiple transformations. MAE\nobreakdash--CT increases the EI compared to MAE, showing that the NN lookup acts as a data-driven augmentation. MAE\nobreakdash--CT$_{aug}$ can further increase the EI against color augmentations due to the hand-crafted augmentations used during contrastive tuning.
}
\label{tab:effective_invariance}
\end{table}

\paragraph{Supervised MAE\nobreakdash--CT.} To explore the effect of perfect nearest neighbor retrieval, we apply contrastive tuning to a MAE pre-trained ViT-L/16 with an oracle (MAE\nobreakdash--CT$_{oracle}$) where only samples of the same class are considered as candidates for the nearest neighbor lookup. After 30 epochs of contrastive tuning such a model achieves $83.9\%$ linear probing performance. In Table~\ref{tab:finetune_minaug} we compare it against a fully fine-tuned MAE with varying degrees of input augmentations. With only crop \& flip augmentation, MAE\nobreakdash--CT$_{oracle}$ outperforms full fine-tuning.

We also validate our fine-tuning pipeline by fine-tuning our MAE pre-trained ViT-L/16 according to the fine-tuning protocol of MAE~\cite{he2022masked}. We achieve a similar performance of $85.8\%$ compared to $85.9\%$ reported in MAE.

\begin{table}[h]
\begin{center}
\begin{tabular}{lc}
Method & Accuracy \\
\hline
\multicolumn{2}{l}{\textit{Crop \& flip only}} \\
MAE\nobreakdash--CT$_{oracle}$ linear probe & \textbf{83.9} \\
MAE fine-tuned & 83.1 \\
+ Label smoothing~\cite{szegedy2016rethinkinginception}& 83.7 \\
\hline
\multicolumn{2}{l}{\textit{Additional input augmentations}} \\
+ Mixup~\cite{zhang2018mixup} & 84.5 \\
+ Cutmix~\cite{yun2019cutmix} & 84.8 \\
+ RandomErasing~\cite{zhong2020RandomErasing} & 84.8 \\
+ RandAugment~\cite{nips2020randaug} & \textbf{85.8} \\
\end{tabular}
\end{center}
\caption{ MAE\nobreakdash--CT$_{oracle}$ with linear probing compared to fine-tuning a MAE on ImageNet with a ViT-L/16 as encoder. When only crop \& flip is used, MAE\nobreakdash--CT$_{oracle}$ performs better with linear probing than a fully fine-tuned MAE. Rows starting with "+" add an additional component to the setting of the previous row. }
\label{tab:finetune_minaug}
\end{table}

\paragraph{Extended results.} Table~\ref{tab:results_ext} on the next page presents an extended version of the linear probing and $k$-NN result tables of the main paper. We include results from models that use smaller patches or a higher input resolution.

\begin{table*}[t]
\begin{center}
\begin{tabular}{lcccccccc}
Method    & Architecture & Epochs & GV/LV/T & Mask ratio & Probe & $k$-NN   \\
\hline
\multicolumn{4}{l}{\textit{minimal image augmentations}} \\
SimMIM~\cite{xie2022simmim} & ViT-B/16 & 800 & 1 & 0.5 & 56.7 & - \\ 
MAE~\cite{he2022masked}  & ViT-B/16  & 1600 & 1  & 0.75 & 68.0 & 50.0 \\
MAE~\cite{he2022masked}  & ViT-L/16  & 1600 & 1  & 0.75 & 76.0 & 60.6 \\
MAE~\cite{he2022masked}   & ViT-H/14  & 1600 & 1  & 0.75 & 77.2 & 58.9 \\
MAE (reimpl.)   & ViT-B/16  & 800 & 2 & 0.75 & 66.7 & 51.1 \\
MAE (reimpl.)    & ViT-L/16  & 800 & 2 &  0.75 & 75.9 & 57.6 \\
MAE (reimpl.)    & ViT-H/16  & 800 & 2 &  0.75 & 78.0 & 61.1 \\
I-JEPA~\cite{assran2023self}   & ViT-B/16  & 600 & 1* / 4* / 1 & * & 72.9 & - \\
I-JEPA~\cite{assran2023self}   & ViT-L/16  & 600 & 1* / 4* / 1 & * & 77.5 & - \\
I-JEPA~\cite{assran2023self}   & ViT-H/14  & 300 & 1* / 4* / 1 & * & 79.3 & 71.6 \\
I-JEPA~\cite{assran2023self}   & ViT-H/16$_{448}$  & 300 & 1* / 4*  / 1 & * & 81.1 & 70.1 \\
MAE\nobreakdash--CT$_{min}$   & ViT-B/16  & 800; 20 & 2 & 0.75; 0 &  73.5 & 64.1 \\
MAE\nobreakdash--CT$_{min}$    & ViT-L/16  & 800; 20 & 2  & 0.75; 0 & 80.2 & 78.0 \\
MAE\nobreakdash--CT$_{min}$    & ViT-H/16  & 800; 30 & 2  & 0.75; 0 & \textbf{81.5} & \textbf{79.4} \\
\hline
\multicolumn{4}{l}{\textit{extensive image augmentations}} \\
NNCLR~\cite{dwibedi2021} & ViT-B/16 & 1000 & 2 & - & 76.5 & -\\
MoCo v3~\cite{chen2021mocov3} & ViT-B/16  & 600 & 2 & - & 76.7 & 72.6 \\
MoCo v3~\cite{chen2021mocov3}   & ViT-L/16  &  300 & 2  & - & 77.6 & - \\
MoCo v3~\cite{chen2021mocov3} & ViT-L/7-BN & 300 & 2  & - & 81.0 & - \\
MoCo v3~\cite{chen2021mocov3}   & ViT-H/14  & 300 & 2  & - & 78.1 &  -\\
MoCo v3~\cite{chen2021mocov3}   & ViT-H/14-BN  & 300 & 2  & - & 79.1 & - \\
DINO~\cite{caron2021emerging}    & ViT-S/8  & 400 & 2 / 10 / 2 & 79.7 & 78.3 \\
DINO~\cite{caron2021emerging}     & ViT-B/16  & 400 & 2 / 10 / 2 & 78.2 & 76.1 \\
DINO~\cite{caron2021emerging}     & ViT-B/8  & 400 & 2 / 10 / 2 & -  & 80.1 & 77.4  \\
iBOT~\cite{zhou2021image} & ViT-B/16  & 400 & 2 / 10 / 2 & $\approx$ 0.15$^\dag$ & 79.5 & 77.1 \\
iBOT~\cite{zhou2021image}   & ViT-L/16  & 250 & 2 / 10 / 2 & $\approx$ 0.15$^\dag$ & 81.0 & 78.0 \\
MSN~\cite{assran2022masked}  & ViT-L/7  & 200  & 1 / 10 / 1 & 0.7 & 80.7 & -  \\
CMAE~\cite{huang2022contrastive} & ViT-B/16 & 1600 & 1 / - / 1 & 0.75 & 73.9 & -  \\
Layer Grafting~\cite{jiang2023layergrafting}   & ViT-B/16  & 1600; 300 & 1; 2 / - / 
- ; 2 & 0.75; 0  & 77.9 & 75.4 \\
Layer 
Grafting~\cite{jiang2023layergrafting}    & ViT-L/16  & 1600; 300 & 1; 2 / - / - ; 2 & 0.75; 0 & 81.0 & 77.3 \\
MAE\nobreakdash--CT$_{aug}$   & ViT-B/16  & 800; 80 & 2  & 0.75; 0 & 76.9 & 73.4 \\
MAE\nobreakdash--CT$_{aug}$    & ViT-L/16  & 800; 40 & 2  & 0.75; 0 & 81.5 & 79.1 \\
MAE\nobreakdash--CT$_{aug}$     & ViT-H/16  & 800; 40 & 2  & 0.75; 0 & \textbf{82.2} & \textbf{79.8} \\
\\
\end{tabular}
\caption{Comparison of linear probing and $k$-NN accuracies, including results of methods that use smaller patches or higher resolutions. 
Furthermore, training details are listed. Namely, the training duration, the number of global views (GV), the number of local views (LV, multi-crop), the number of teacher forward passes (T) and the masking ratio. \\ 
*~I-JEPA uses a multiblock-masking scheme,  consisting of one context block and four target blocks (no encoder forward pass) per sample. Thus the effective masking ratio is a result of the size of the context block, the sizes of the target blocks and the overlap between them. \\
\dag~iBOT applies masking only half of the time during training. If masking is applied the masking ratio is sampled between 0.1 and 0.5.  \\
}
\label{tab:results_ext}
\end{center}
\end{table*}

\newpage
\section{Cluster Analysis}
\label{app:cluster_analysis}

\paragraph{Experiment details.} The clustering performance metrics in Table 4 of the main paper are obtained by applying $k$-means to the [CLS] token in the last layer of each ViT encoder. Only for MAE\nobreakdash--CT and MAE\nobreakdash--CT$_{aug}$ with ViT-B we use the average over the patch tokens of the last encoder layer for clustering. For each method we encoded the validation set of ImageNet and standardized the embedding using its mean and standard deviation. The standardized embedding is then clustered 100 times with mini-batch $k$-means\footnote{We used the publicly available implementation of scikit-learn \cite{scikit-learn}, see \url{https://scikit-learn.org/stable/modules/generated/sklearn.cluster.MiniBatchKMeans.html}.}, a mini-batch variant of $k$-means that scales better with larger data sets \cite{Sculley10}.  
From the 100 runs we selected the one with the lowest $k$-means loss and calculated the cluster accuracy \cite{YangXNYZ10_cluster_accuracy} w.r.t. the 1000 ground truth ImageNet classes. For ImageNet-Dogs15~\cite{ChangWMXP17_imagenetdogs} we performed the same procedure with the number of clusters $k=15$.

\paragraph{Quantitative results.} In Table \ref{tab:clustering_metrics_additional_best} we show for completeness the best (w.r.t.\ the lowest $k$-means loss) clustering performance for ImageNet and in Table \ref{tab:clustering_metrics_additional_average} we report the average results. For ImageNet-Dogs15 we report the best results in Table \ref{tab:clustering_metrics_dogs_best} and average results in Table \ref{tab:clustering_metrics_dogs_average}. In addition to the cluster accuracy (ACC) results presented in the main paper we report the commonly used clustering metrics Normalized Mutual Information (NMI; \cite{NMI}), Adjusted Mutual Information (AMI; \cite{NMI}) and Adjusted Rand Index (ARI; \cite{ARI}) as well. ACC, NMI, AMI and ARI are all multiplied by 100, so they range between 0 and 100, where higher values indicate a better match with the ground truth labels. The ACC standard deviations over the 100 runs are small for ImageNet, ranging from 0.2\% to 0.7\% and higher for ImageNet-Dogs15, ranging from 0.6\% to 4.3\%.

\begin{figure}[ht!]
\begin{center}
\begin{overpic}[width=\linewidth]{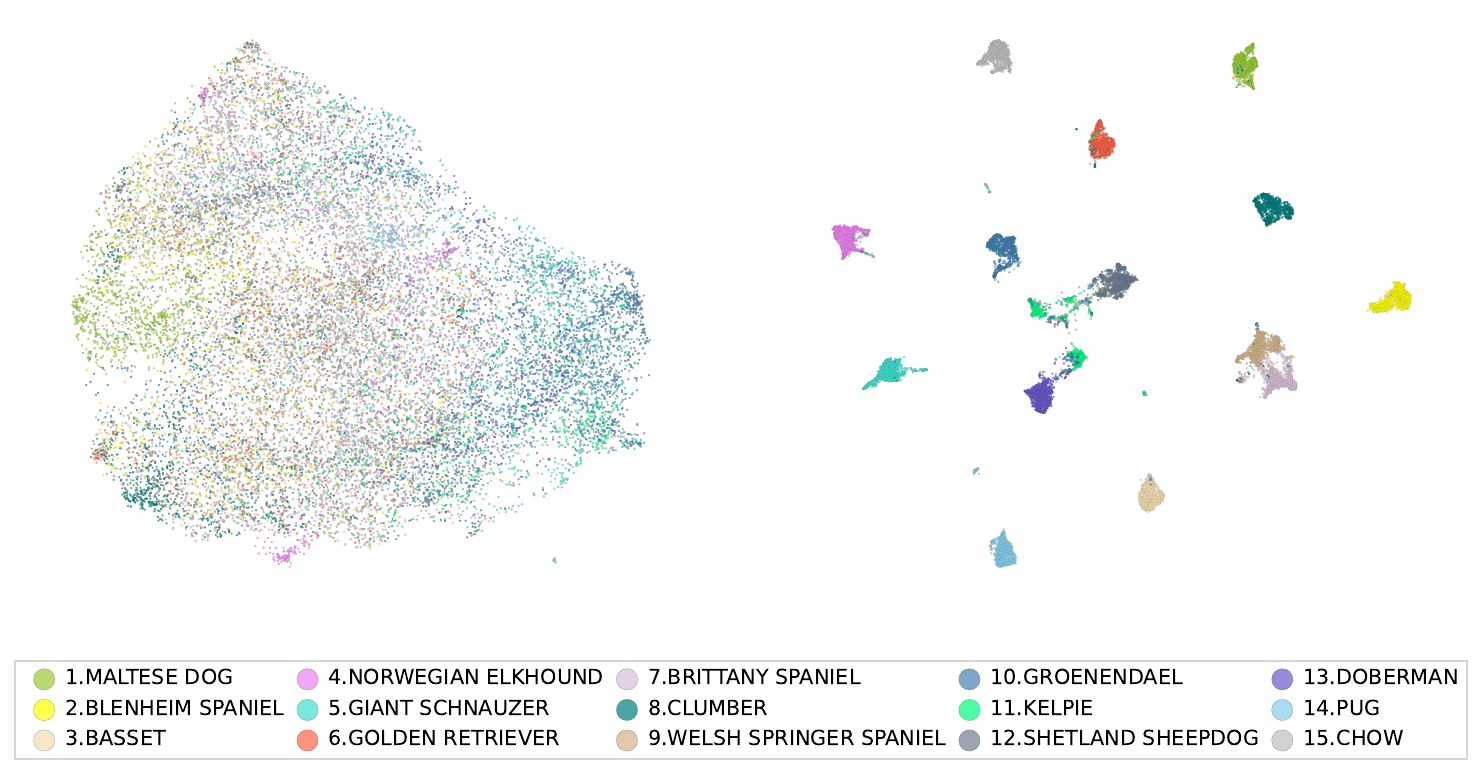}
\begin{scriptsize}
\put(30,120){MAE: 18.7\%}
\put(130,120){MAE\nobreakdash--CT$_{min}$: 94.3\%}
\end{scriptsize}
\end{overpic}
\end{center}
\caption{Visualization of UMAP embeddings from the [CLS] token of ViT-H/16 over MAE and MAE\nobreakdash--CT$_{min}$ with corresponding $k$-means cluster accuracies (w.r.t.\ lowest $k$\nobreakdash-means loss) for ImageNet-Dogs15. MAE\nobreakdash--CT$_{min}$ clearly improves the separation of the 15 classes.
}
\label{fig:dogs_umap_huge}
\end{figure}

\begin{figure}[t!]
\begin{center}
\includegraphics[width=\linewidth]{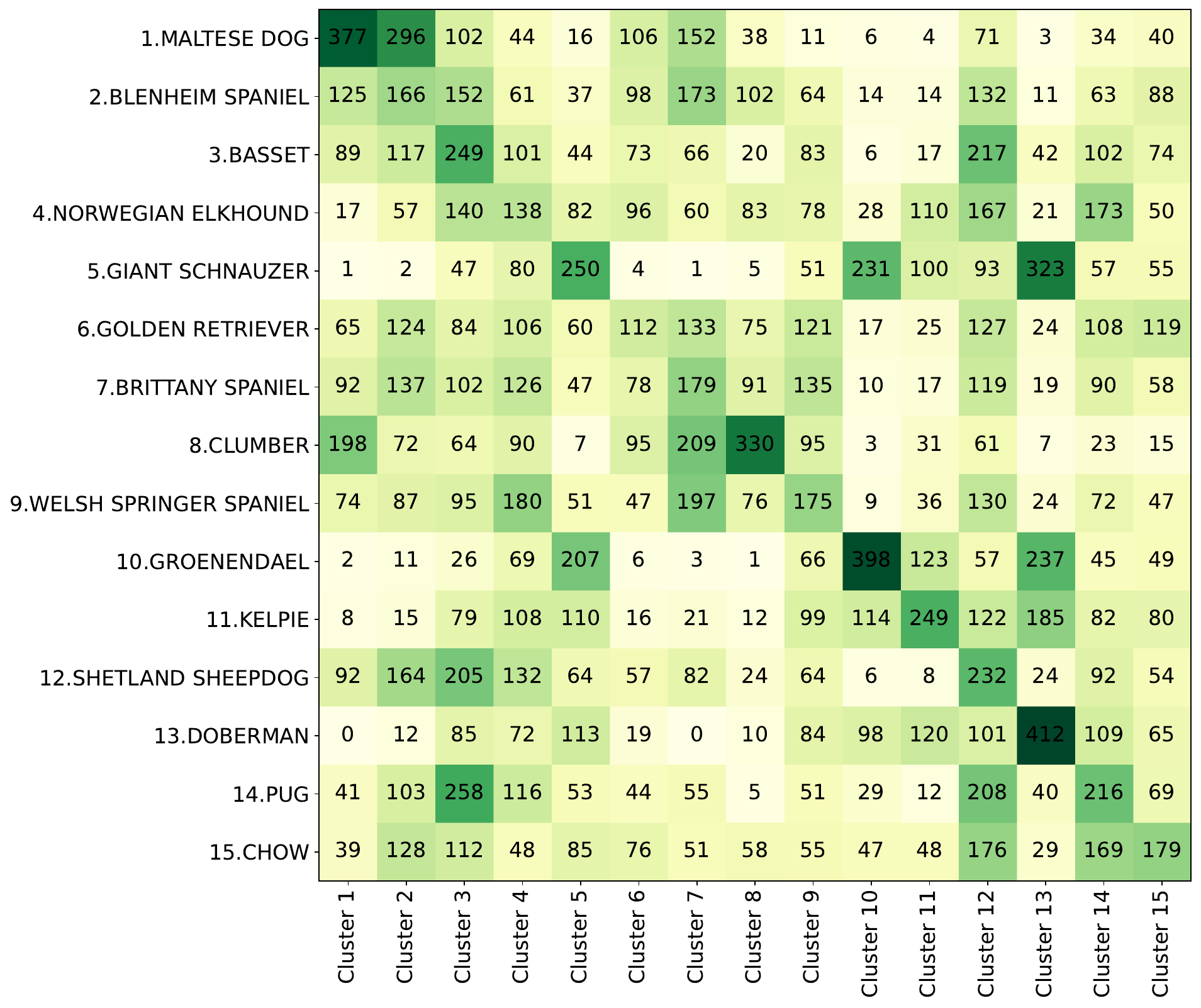}
\includegraphics[width=\linewidth]{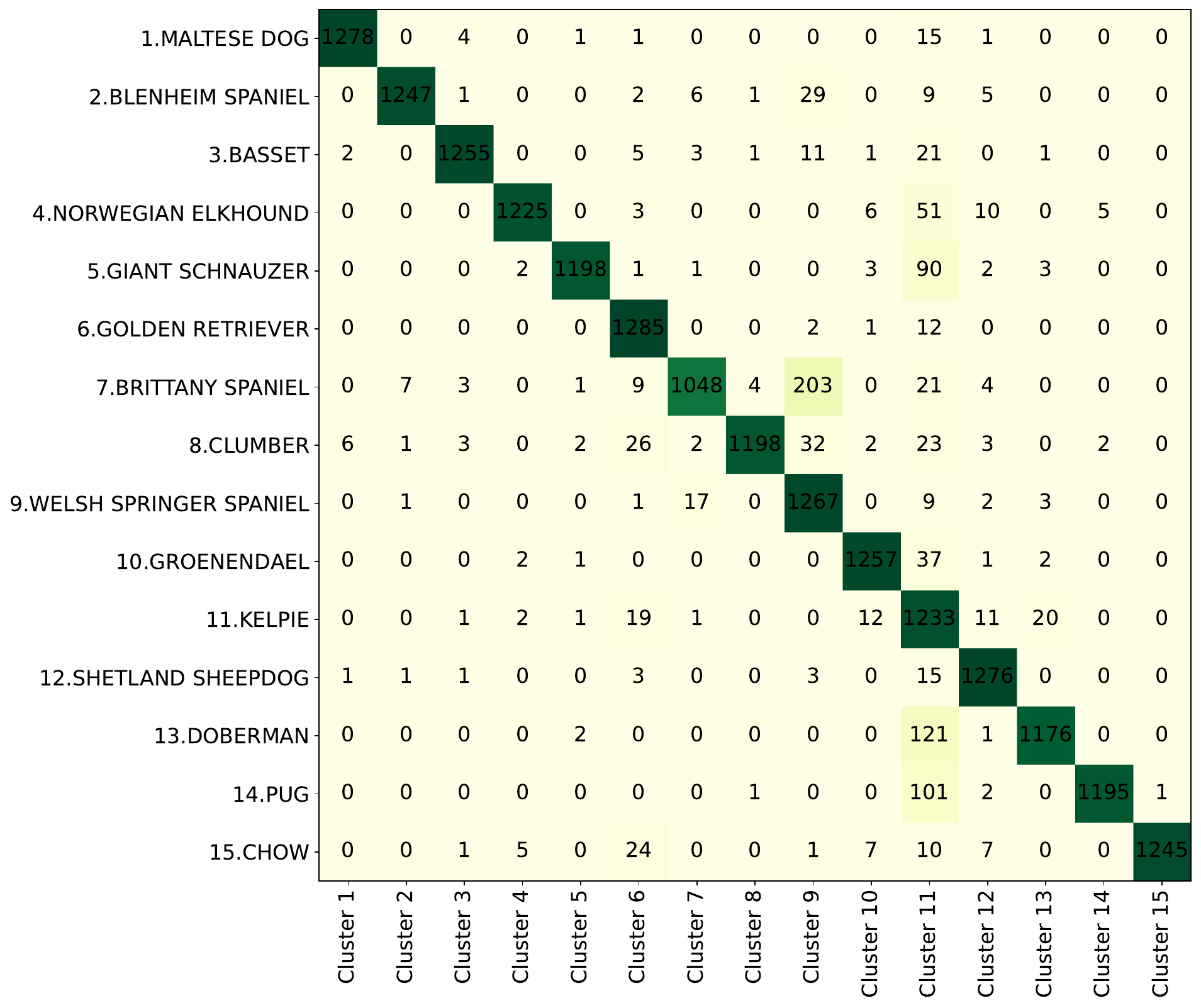}
\end{center}
   \caption{Confusion matrices w.r.t. the clustering result with lowest $k$-means loss for MAE (upper) with 18.7\% cluster accuracy and MAE\nobreakdash--CT$_{min}$ (lower) with 94.3\% for ViT-H/16. MAE\nobreakdash--CT$_{min}$ discovers clusters that correspond well to the ground truth dog breeds.}
\label{fig:dogs_confusion_matrices}
\end{figure}

\begin{figure*}[t!]
\begin{center}
\begin{overpic}[width=0.485\linewidth]{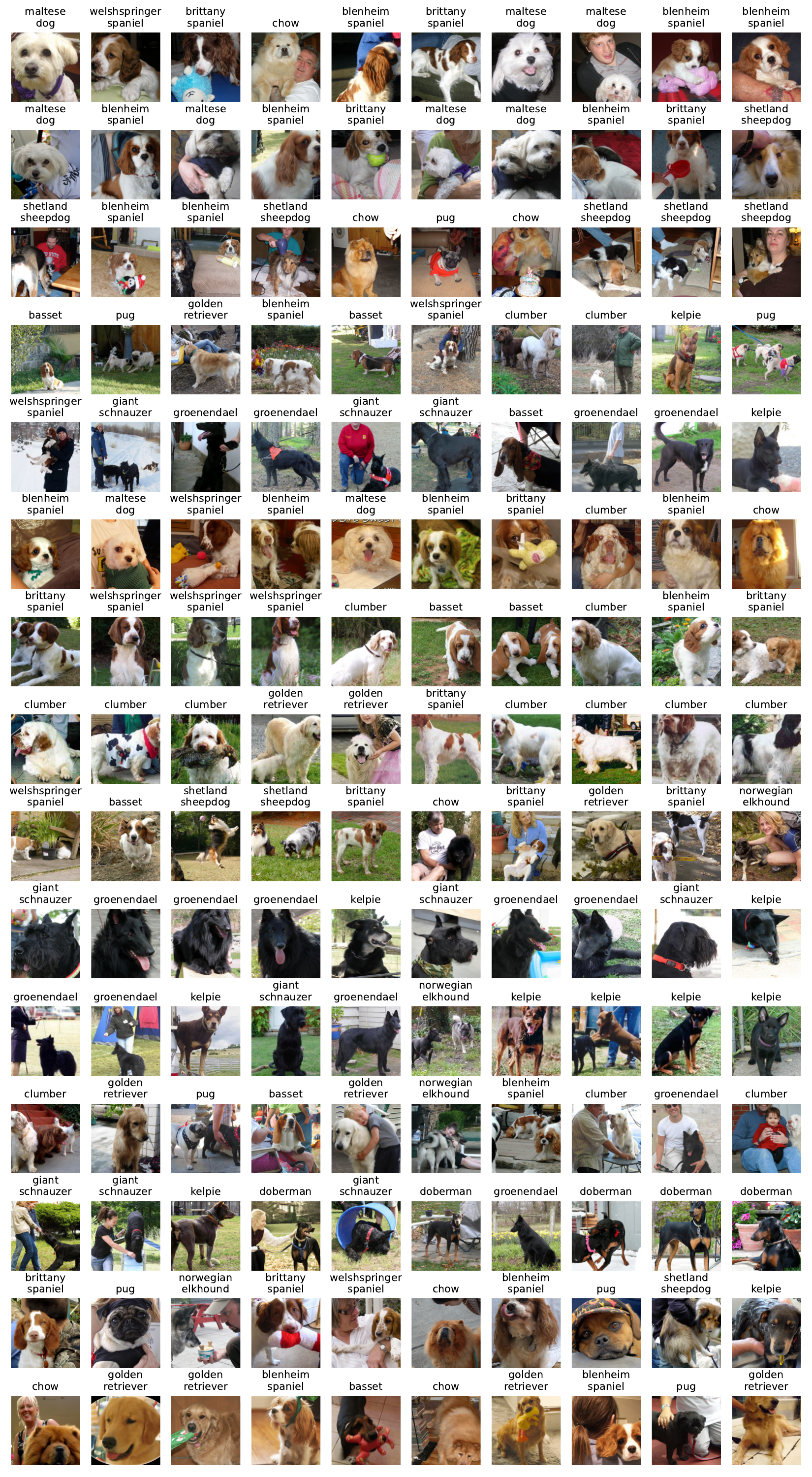}
\begin{scriptsize}
\put(248,421){C1}
\put(248,392){C2}
\put(248,363){C3}
\put(248,334){C4}
\put(248,305){C5}
\put(248,276){C6}
\put(248,247){C7}
\put(248,218){C8}
\put(248,189){C9}
\put(248,160){C10}
\put(248,131){C11}
\put(248,98){C12}
\put(248,72){C13}
\put(248,42){C14}
\put(248,15){C15}
\end{scriptsize}
\end{overpic}
\hfill
\includegraphics[width=0.485\linewidth]{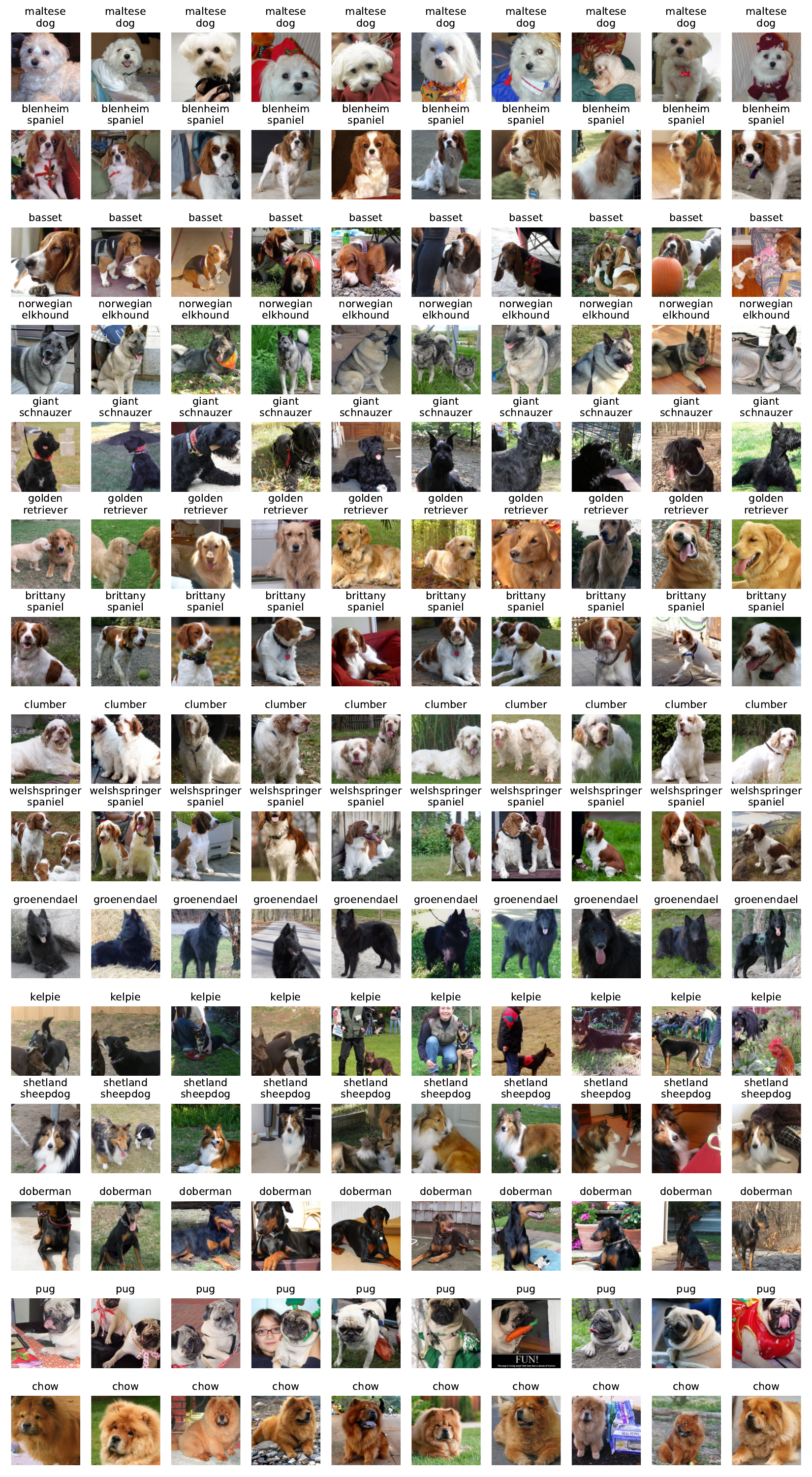}

\end{center}
   \caption{\textbf{Cluster retrieval:} Ten nearest neighbors per $k$-means cluster center for MAE (left) and MAE\nobreakdash--CT$_{min}$ (right). Each row corresponds to one cluster (C1 to C15) found in the [CLS] token embedding of ViT-H/16 for Imagenet-Dogs15. MAE clusters correspond more to background and surface characteristics like color, while MAE\nobreakdash--CT$_{min}$ found clusters that correspond to the 15 dog breeds. The  mapping from each cluster to dog breed for MAE\nobreakdash--CT$_{min}$ can be found in the confusion matrix in Figure \ref{fig:dogs_confusion_matrices}, e.g.\ the first row is \textit{cluster} $1$ (C1) and corresponds to the class \textit{maltese dog}. Additionally, the ground truth label is shown on top of each image (best viewed with zooming in).}
\label{fig:cluster_retrievals_detailed}
\end{figure*}

\begin{figure*}[htp]
\begin{center}
\includegraphics[clip,width=\textwidth]{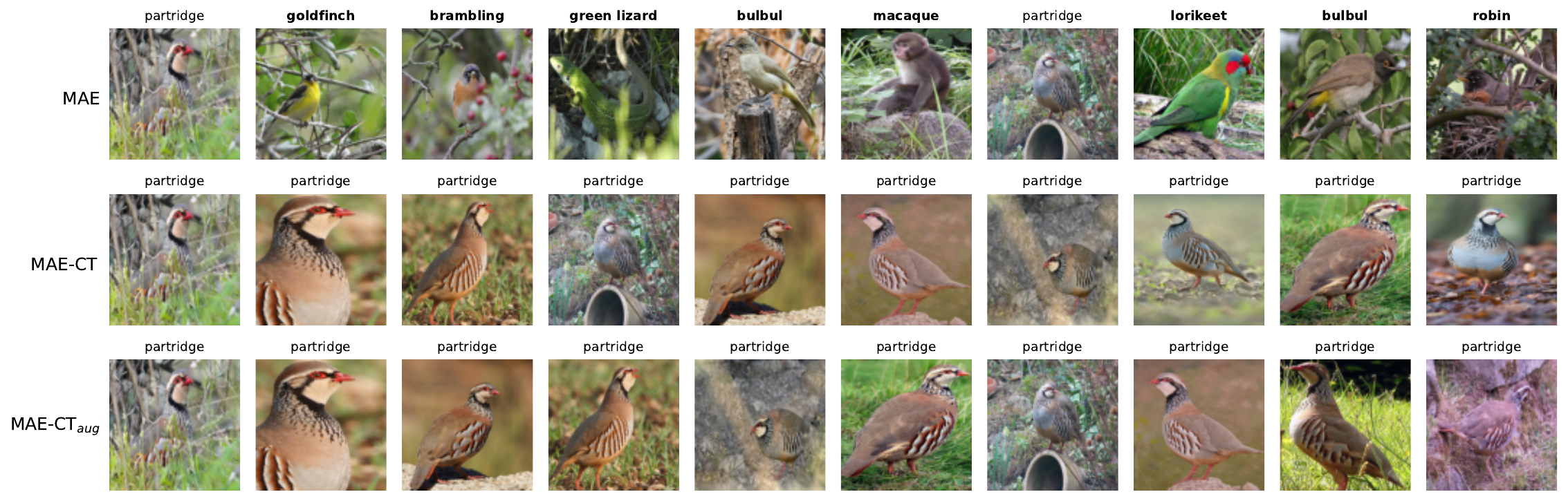} 
\hfill
\includegraphics[clip,width=\textwidth]{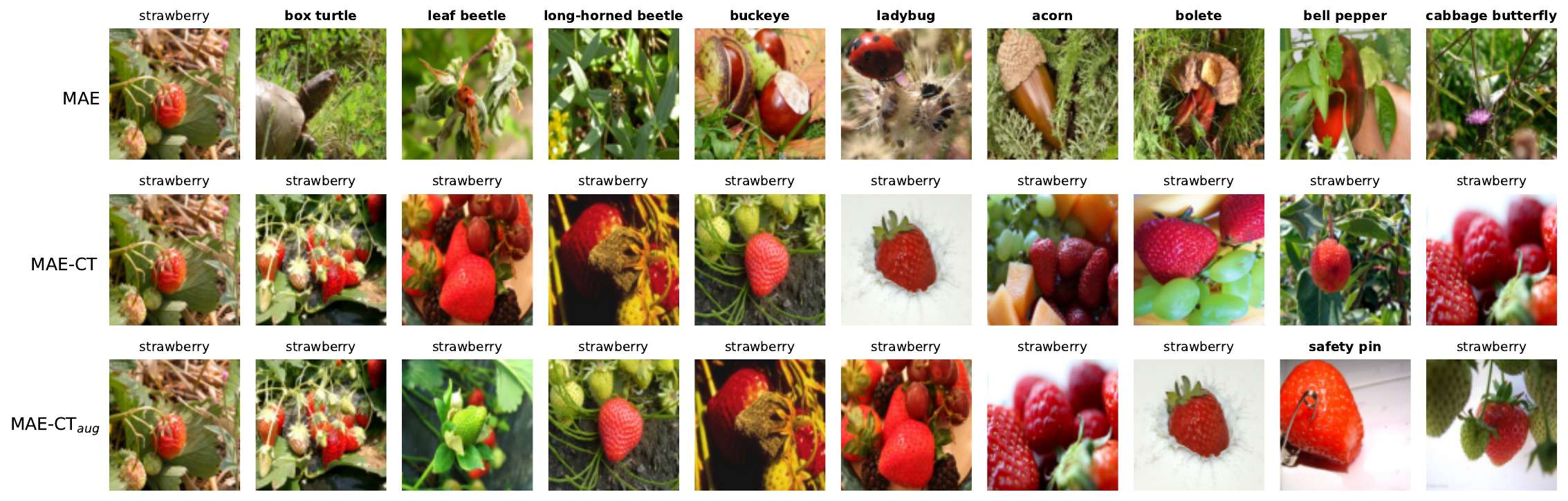}
\hfill
\includegraphics[clip,width=\textwidth]{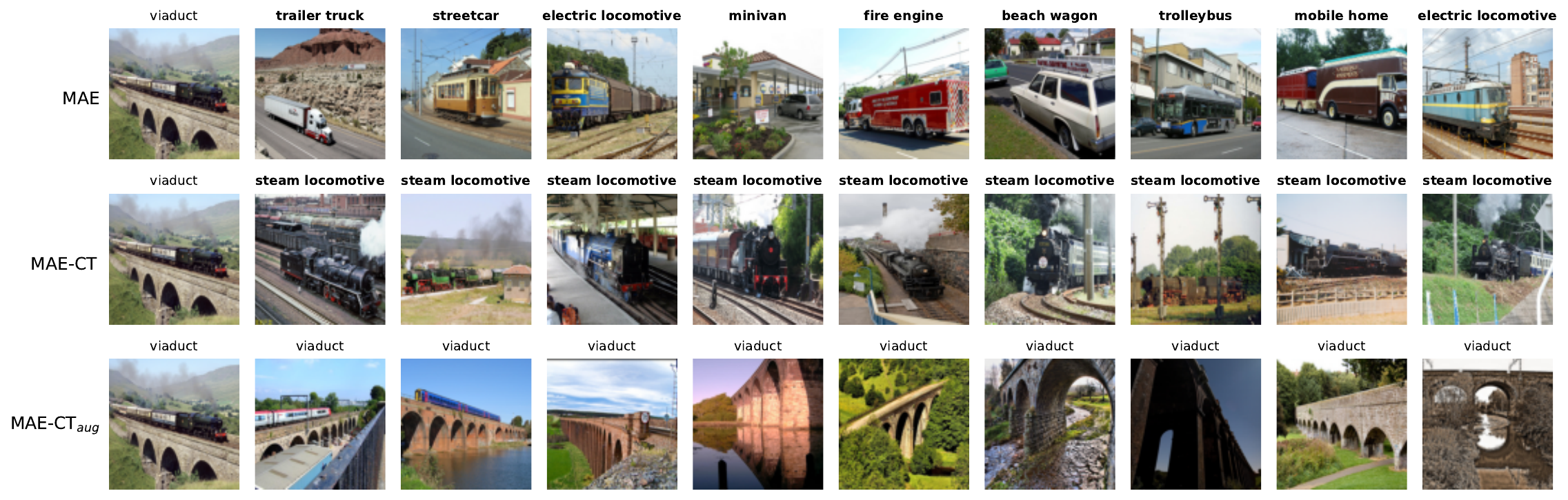}
\end{center}
\caption{\textbf{Nearest neighbor retrieval:} Nearest neighbors for three different sample images (first column) for MAE, MAE\nobreakdash--CT and MAE\nobreakdash--CT$_{aug}$. 
The NNs are computed by calculating the [CLS] token embedding of a ViT-H for all images in the ImageNet validation set and searching for the most similar embeddings in terms of cosine similarity.
The first and the second example images (\textit{partridge} and \textit{strawberry}) show that the NNs of the MAE are similar in the context in which the image was taken but almost never contain the correct class.
However, the NNs of the MAE\nobreakdash--CT and the MAE\nobreakdash--CT$_{aug}$ are images of the same class as the example images.
The third sample image is labeled as \textit{viaduct}, 
but the image clearly also contains a \textit{steam locomotive}. 
The NNs of the MAE all belong to different classes, 
which shows that the MAE has not captured the semantics of a \textit{viaduct} or a \textit{steam locomotive}.
The NNs of the MAE\nobreakdash--CT and the MAE\nobreakdash--CT$_{aug}$ are all of the same class, 
but it is either the \textit{steam locomotive} or the \textit{viaduct}. 
All NNs of MAE\nobreakdash--CT$_{aug}$ are images of other \textit{viaducts} recognizable by their characteristic shape,
showing that the augmentations of MAE\nobreakdash--CT$_{aug}$ result in a representation that focuses more on shapes and less on characteristic colors, like black locomotives.
}
\label{fig:nearest_neighbors_examples}
\end{figure*}

\paragraph{Qualitative results.} In Figure \ref{fig:dogs_umap_huge} we show the UMAP \cite{mcinnes2018umap-software} plots for the [CLS] token embedding of ImageNet-Dogs15 of MAE vs. MAE\nobreakdash--CT$_{min}$ for ViT-H/16. We see that the cluster structure is much more visible for MAE\nobreakdash--CT$_{min}$. Figure \ref{fig:dogs_confusion_matrices} shows the corresponding confusion matrices, where clusters are matched to the ground truth classes using a variant of the Hungarian algorithm \cite{scipy_hungarian_algo_Crouse16}.  
In Figure \ref{fig:cluster_retrievals_detailed} we show the corresponding cluster retrievals for MAE vs. MAE\nobreakdash--CT$_{min}$ where we can see again that the nearest neighbors of $k$-means cluster centers discovered by MAE\nobreakdash--CT$_{min}$ are focusing more on the different dog breeds than on the background as in MAE.
Additionally,  Figure~\ref{fig:nearest_neighbors_examples} shows three samples of ImageNet and the nearest neighbors for MAE, MAE\nobreakdash--CT$_{min}$ and MAE\nobreakdash--CT$_{aug}$, respectively.

\vspace{-0.2pt}

\newpage

\begin{table*}
\begin{center}
\begin{tabular}{l|cccc|cccc|cccc}
        & \multicolumn{4}{c|}{{ViT-B/16}} & \multicolumn{4}{c|}{{ViT-L/16}} & \multicolumn{4}{c}{{ViT-H/16}} \\
Method                                 & ACC           & NMI           & AMI           & ARI           & ACC           & NMI           & AMI           & ARI           & ACC           & NMI           & AMI           & ARI           \\
\hline
\textit{minimal image augmentations} & & & & & & & & & & & & \\

MAE~(reimpl.)                    & 13.8          & 51.4          & 20.7          & 4.4           & 14.3          & 51.9          & 22.8          & 4.6           & 11.1          & 48.2          & 17.1          & 2.9           \\
MAE\nobreakdash--CT$_{min}$                & \textbf{35.3} & \textbf{68.2} & \textbf{45.8} & \textbf{19.7} & \textbf{54.9} & \textbf{80.6} & \textbf{66.8} & \textbf{34.5} & \textbf{58.0} & \textbf{81.8} & \textbf{69.3} & \textbf{36.8} \\
\hline
\textit{extensive image augmentations} & & & & & & & & & & & & \\
Moco v3 & 43.0          & 73.0          & 55.5          & 23.1          & -             & -             & -             & -             & -             & -             & -             & -             \\
DINO    & 48.0          & 77.9          & 63.2          & 31.2          & -             & -             & -             & -             & -             & -             & -             & -             \\
MSN      & \textbf{54.2} & \textbf{79.9} & \textbf{66.6} & 30.2          & 45.4          & 76.2          & 63.2          & 19.5          & -             & -             & -             & -             \\
iBOT        & 50.0          & 79.1          & 64.9          & \textbf{32.6} & 52.0          & 80.5          & 66.8          & 35.0          & -             & -             & -             & -             \\
MAE\nobreakdash--CT$_{aug}$                   & 46.2          & 76.0          & 59.0          & 29.8          & \textbf{56.9} & \textbf{81.7} & \textbf{68.4} & \textbf{40.5} & 54.8          & 81.1          & 67.6          & 38.9          \\
\end{tabular}
\end{center}
\caption{Best cluster performance of $k$-means (w.r.t.\ lowest $k$-means loss) over 100 runs for ImageNet validation set.}
\label{tab:clustering_metrics_additional_best}
\end{table*}

\begin{table*}
\begin{center}
\begin{tabular}{l|cccc|cccc|cccc}
        & \multicolumn{4}{c|}{{ViT-B/16}} & \multicolumn{4}{c|}{{ViT-L/16}} & \multicolumn{4}{c}{{ViT-H/16}} \\
Method                                 & ACC           & NMI           & AMI           & ARI           & ACC           & NMI           & AMI           & ARI           & ACC           & NMI           & AMI           & ARI           \\
\hline
\textit{minimal image augmentations} & & & & & & & & & & & & \\
MAE~(reimpl.)                    & 13.5          & 51.0          & 20.8          & 4.3           & 13.9          & 51.2          & 22.9          & 4.5           & 10.6          & 47.5          & 17.0          & 2.7           \\
MAE\nobreakdash--CT$_{min}$                           & \textbf{35.0} & \textbf{68.0} & \textbf{45.4} & \textbf{19.5} & \textbf{54.1} & \textbf{80.6} & \textbf{66.9} & \textbf{34.7} & \textbf{57.1} & \textbf{81.7} & \textbf{69.2} & \textbf{35.6} \\
\hline
\textit{extensive image augmentations} & & & & & & & & & & & & \\
Moco v3 & 41.5          & 72.4          & 55.2          & 21.2          & -             & -             & -             & -             & -             & -             & -             & -             \\
DINO    & 46.5          & 77.3          & 62.7          & 29.0          & -             & -             & -             & -             & -             & -             & -             & -             \\
MSN      & \textbf{52.7} & \textbf{79.4} & \textbf{65.9} & 28.8          & 43.6          & 75.6          & 62.7          & 18.6          & -             & -             & -             & -             \\
iBOT        & 49.2          & 78.7          & 64.6          & \textbf{31.6} & 51.2          & 80.1          & 66.4          & 34.1          & -             & -             & -             & -             \\
MAE\nobreakdash--CT$_{aug}$                   & 45.3          & 75.7          & 58.8          & 28.6          & \textbf{55.6} & \textbf{81.4} & \textbf{68.0} & \textbf{39.4} & 54.7          & 80.8          & 67.3          & 37.8          \\
\end{tabular}
\end{center}
\caption{Average cluster performance of $k$-means over 100 runs for ImageNet validation set.}
\label{tab:clustering_metrics_additional_average}
\end{table*}

\begin{table*}
\begin{center}
\begin{tabular}{l|cccc|cccc|cccc}
        & \multicolumn{4}{c|}{{ViT-B/16}} & \multicolumn{4}{c|}{{ViT-L/16}} & \multicolumn{4}{c}{{ViT-H/16}} \\
Method                                 & ACC           & NMI           & AMI           & ARI           & ACC           & NMI           & AMI           & ARI           & ACC           & NMI           & AMI           & ARI           \\
\hline
\textit{minimal image augmentations} & & & & & & & & & & & & \\
MAE~(reimpl.)                    & 16.0          & 8.1           & 7.9           & 3.2           & 18.9          & 12.8          & 12.6          & 5.2           & 18.7          & 11.7          & 11.6          & 4.7           \\
MAE\nobreakdash--CT$_{min}$                           & \textbf{70.8} & \textbf{69.1} & \textbf{69.0} & \textbf{54.3} & \textbf{84.2} & \textbf{85.6} & \textbf{85.6} & \textbf{76.4} & \textbf{94.3} & \textbf{90.4} & \textbf{90.4} & \textbf{87.9} \\
\hline
\textit{extensive image augmentations} & & & & & & & & & & & & \\
Moco v3 & 72.7          & 71.9          & 71.8          & 59.3          & -             & -             & -             & -             & -             & -             & -             & -             \\
DINO    & 79.0          & 78.9          & 78.9          & 69.4          & -             & -             & -             & -             & -             & -             & -             & -             \\
MSN      & \textbf{81.0} & \textbf{83.6} & \textbf{83.6} & \textbf{73.7} & 83.3          & 84.2          & 84.2          & 75.9          & -             & -             & -             & -             \\
iBOT        & 77.2          & 78.8          & 78.8          & 68.5          & 69.4          & 72.8          & 72.8          & 59.9          & -             & -             & -             & -             \\
MAE\nobreakdash--CT$_{aug}$                   & 78.1          & 77.4          & 77.3          & 65.0          & \textbf{84.8} & \textbf{87.8} & \textbf{87.8} & \textbf{81.1} & 83.7          & 86.5          & 86.4          & 79.3          \\
\end{tabular}
\end{center}
\caption{Best cluster performance of $k$-means (w.r.t.\ lowest $k$-means loss) over 100 runs for ImageNet-Dogs15.}
\label{tab:clustering_metrics_dogs_best}
\end{table*}

\begin{table*}
\begin{center}
\begin{tabular}{l|cccc|cccc|cccc}
        & \multicolumn{4}{c|}{{ViT-B/16}} & \multicolumn{4}{c|}{{ViT-L/16}} & \multicolumn{4}{c}{{ViT-H/16}} \\
Method                                 & ACC           & NMI           & AMI           & ARI           & ACC           & NMI           & AMI           & ARI           & ACC           & NMI           & AMI           & ARI           \\
\hline
\textit{minimal image augmentations} & & & & & & & & & & & & \\
MAE~(reimpl.)                    & 15.7          & 8.5           & 8.3           & 3.3           & 18.5          & 12.3          & 12.1          & 5.2           & 17.6          & 11.7          & 11.5          & 4.8           \\
MAE\nobreakdash--CT$_{min}$                           & \textbf{67.9} & \textbf{69.2} & \textbf{69.2} & \textbf{53.5} & \textbf{79.9} & \textbf{83.4} & \textbf{83.4} & \textbf{72.7} & \textbf{87.4} & \textbf{88.2} & \textbf{88.2} & \textbf{82.1} \\
\hline
\textit{extensive image augmentations} & & & & & & & & & & & & \\
Moco v3 & 66.7          & 68.8          & 68.8          & 54.2          & -             & -             & -             & -             & -             & -             & -             & -             \\
DINO    & 71.4          & 75.2          & 75.2          & 62.2          & -             & -             & -             & -             & -             & -             & -             & -             \\
MSN      & \textbf{79.7} & \textbf{82.4} & \textbf{82.3} & \textbf{72.2} & 79.5          & 82.9          & 82.9          & 72.5          & -             & -             & -             & -             \\
iBOT       & 75.2          & 78.6          & 78.6          & 66.9          & 64.8          & 70.5          & 70.4          & 55.4          & -             & -             & -             & -             \\
MAE\nobreakdash--CT$_{aug}$                   & 71.7          & 74.7          & 74.7          & 60.3          & \textbf{84.7} & \textbf{86.5} & \textbf{86.4} & \textbf{79.1} & 83.2          & 86.0          & 86.0          & 78.2          \\
\end{tabular}
\end{center}
\caption{Average cluster performance of $k$-means over 100 runs for ImageNet-Dogs15.}
\label{tab:clustering_metrics_dogs_average}
\end{table*}

\end{document}